\documentclass[pdflatex,sn-mathphys-num]{sn-jnl}% Math and Physical Sciences Numbered Reference Style

\usepackage{threeparttable}
\usepackage{textgreek}
\usepackage{graphicx}%
\usepackage{multirow}%
\usepackage{amsmath,amssymb,amsfonts}%
\usepackage{amsthm}%
\usepackage{mathrsfs}%

\usepackage[title]{appendix}%
\usepackage{xcolor}%
\usepackage{textcomp}%
\usepackage{manyfoot}%
\usepackage{longtable}
\usepackage{booktabs}%
\usepackage{algorithm}%
\usepackage{algorithmicx}%
\usepackage{algpseudocode}%
\usepackage{listings}%
\theoremstyle{thmstyleone}%
\theoremstyle{thmstyletwo}%

\theoremstyle{thmstylethree}%

\begin{document}

\title[Article Title]{LegalEval-Q: A New Benchmark for The Quality Evaluation of LLM-Generated Legal Text}

%%=============================================================%%
%% GivenName	-> \fnm{Joergen W.}
%% Particle	-> \spfx{van der} -> surname prefix
%% FamilyName	-> \sur{Ploeg}
%% Suffix	-> \sfx{IV}
%% \author*[1,2]{\fnm{Joergen W.} \spfx{van der} \sur{Ploeg} 
%%  \sfx{IV}}\email{iauthor@gmail.com}
%%=============================================================%%

\author[1,2]{\fnm{Yunhan} \sur{Li}}\email{D24092110205@cityu.edu.mo}

\author*[1]{\fnm{Gengshen} \sur{Wu}}\email{gswu@cityu.edu.mo}
%\equalcont{These authors contributed equally to this work.}

\affil[1]{\orgdiv{Faculty of Data Science}, \orgname{City University of Macau}, \orgaddress{\street{Avenida Padre Tomás Pereira Taipa}, \city{Macao}, \postcode{999078}, \state{Macao}, \country{China}}}

\affil[2]{\orgdiv{Shenzhen Institutes of Advanced Technology}, \orgname{Chinese Academy of Sciences}, \orgaddress{\street{Xueyuan Avenue}, \city{Shenzhen}, \postcode{518071}, \state{Guangdong}, \country{China}}}

%%==================================%%
%% Sample for unstructured abstract %%
%%==================================%%

\abstract{As large language models (LLMs) are increasingly used in legal applications, current evaluation benchmarks focus mainly on factual accuracy while neglecting important linguistic aspects such as clarity, coherence, and terminology. To address this gap, we first develop a regression-based framework to evaluate legal text quality, second construct a specialized set of legal questions, and third analyze 49 LLMs using this framework. Our study primarily focuses on Chinese legal texts due to data availability, while the methodology itself remains language-agnostic and adaptable to other domains. We identify three key findings: (1) legal text quality plateaus at relatively small scales, with Qwen2.5 models flattening beyond 7B (72B adds only 2.7\%) and Qwen3 models showing an early plateau at 1.7B; (2) engineering choices such as quantization and context length have no statistically significant effect on legal text quality ($p > 0.0167$), supporting cost-efficient deployment; (3) reasoning models consistently outperform base architectures. A significant outcome of our research is the release of a ranking list and trade-off frontier visualization, which highlight the Qwen3 series as the optimal choice for cost–performance trade-offs. This work advances domain-specific evaluation of linguistic quality by integrating multidimensional assessment with data-driven model analysis. We additionally adopt a variance-penalized metric, AdjScore, to robustly assess model performance. Code and models are available at: \url{https://github.com/lyxx3rd/LegalEval-Q}.}


\maketitle

\section{Introduction}

The rapid advancement of LLMs has significantly transformed natural language processing(NLP), with their widespread deployments across specialized domains such as finance \cite{lu2023bbt,xie2024finben,zhang2023fineval}, healthcare \cite{dai2023laiw,fei2309benchmarking,guha2023legalbench}, and legal practice \cite{dada2024clue,wang2023cmb,li2023huatuo}.

Current evaluation benchmarks primarily address three capabilities: domain-specific knowledge mastery \cite{lu2023bbt,dai2023laiw,dada2024clue}, extended context processing \cite{kamradt2023needle,zhao2024longagent}, and reasoning ability \cite{Hendrycks2021Measuring}. However, these benchmarks consistently overlook an essential dimension, namely textual quality. This omission creates a paradox in which large language models may produce factually correct outputs that nevertheless display deficiencies in linguistic quality and form, such as awkward phrasing, stylistic inconsistency, or reduced readability. Although the present study concentrates on Chinese legal texts due to data availability constraints, the proposed methodology has been deliberately designed with cross-linguistic adaptability. The framework relies on architecture-agnostic metrics and evaluation protocols that can be extended to other languages when suitable domain-specific corpora are available. Developing robust methodologies for textual quality assessment therefore remains a pressing need for both academic research and practical applications, particularly in specialized domains.

In the legal domain, two key challenges remain unresolved. First, current evaluation frameworks lack standardized quantitative metrics for capturing the fine-grained aspects of legal text quality. Second, the connection between model scale and architectural design and the resulting text quality remains poorly understood. These gaps limit effective model selection and optimization in legal applications.

To address these challenges, this study this study bridges the gap between computational evaluation, legal expertise, and linguistic expertise. These research gaps underscore fundamental issues at the intersection of knowledge and information systems and their practical applications. From a knowledge and data engineering perspective, the development of reliable quality assessment metrics requires both domain expertise and methodological innovation, achieved by integrating legal professionals’ understanding of textual nuances with systematic approaches to model evaluation. This study introduces a framework for legal text quality assessment that systematically evaluates clarity, coherence, and terminology through a data-driven approach. By developing quantitative evaluation metrics and conducting extensive model analysis, we establish a comprehensive methodology tailored to the evaluation of LLM-generated legal texts while providing important insights into the effects of model architecture.

Our study makes three primary contributions. First, we introduce a specialized benchmark for evaluating the quality of legal texts. This benchmark incorporates a multidimensional assessment model that captures critical attributes such as clarity, coherence, and precision in terminology. Second, we construct a comprehensive validation set composed of legal questions from diverse subdomains, enabling systematic evaluation of model responses across varied legal contexts. Third, through extensive empirical analysis, we examine how fundamental characteristics of models, in particular their parameter scale and fine-tuning strategies, influence the quality of generated texts. Taken together, our framework offers practitioners practical guidance for selecting language models in legal applications while also establishing standardized evaluation protocols that support future research on domain-specific text quality assessment.

\section{Related Work}

%Traditional NLG evaluation: focusing on the limitations of reasoning capabilities%
\subsection{Traditional Natural Language Generation Evaluation}

With the proliferation of LLMs across diverse domains, establishing rigorous evaluation methodologies has become critical. Current benchmarks predominantly focus on task-oriented or reasoning-based assessments through verifiable instructions. The IFEval framework \cite{zhou2023instructionfollowingevaluationlargelanguage} exemplifies this approach by quantifying instruction-following accuracy via objective metrics. For complex reasoning, the BIG-Bench Hard suite \cite{suzgun-etal-2023-challenging} evaluates multi-step problem-solving, while the MATH dataset \cite{Hendrycks2021Measuring} provides a standardized testbed for mathematical deduction. For domain-specific challenges, benchmarks adopt specialized designs: GPTA \cite{rein2024gpqa}, which curates expert-level questions in STEM fields, and MuSR \cite{sprague2024musrtesting}targets soft reasoning in narrative contexts. Knowledge-intensive evaluations have evolved from MMLU \cite{hendrycks2020measuring} to its advanced variant MMLU-Pro \cite{wang2024mmlupro}, incorporating adversarial noise to better probe reasoning robustness. For non-English contexts, C-Eval \cite{huang2023c} simultaneously measures Chinese linguistic competence and cross-domain knowledge application. While these methods excel at quantifying task completion, such as solution accuracy and instruction adherence, structured reasoning, they reveal fundamental limitations when applied to open-domain text quality assessment.

%Early exploration of text quality assessment: from single dimension to shallow semantics%
\subsection{Early Exploration of Text Quality Assessment}

Current evaluation of generated text remains predominantly reliant on similarity-based metrics \cite{kasai2021bidimensional}, which can be categorized into two paradigms: 1) lexical overlap-based methods such as ROUGE \cite{lin2004rouge} and BLEU \cite{papineni2002bleu} that compute n-gram matches between generated and reference texts, and 2) contextualized embedding-based approaches \cite{zhang2019bertscore, zhao2019moverscore, clark2019sentence} that measure semantic similarity using deep language models.

Despite their prevalent use in more than 60\% of recent natural language generation(NLG) studies \cite{kasai2021bidimensional}, these metrics exhibit fundamental limitations. First, they fail to capture essential quality dimensions including content coherence \cite{reiter2009investigation} and syntactic correctness \cite{stent2005evaluating}. Second, as NLG models advance, surface-level feature matching becomes increasingly inadequate for discriminating system capabilities \cite{gehrmann2023repairing}. Most problematically, this similarity-based paradigm introduces circular dependency on reference texts, inherently constraining the assessment of creative or open-ended generation where optimal outputs may diverge from provided references. Similar limitations have been observed in Q\&A systems, where Albassami et al. \cite{albassami2025comprehensive} note that traditional metrics often fail to capture domain-specific linguistic quality, particularly in professional contexts.

%LLM-driven text quality evaluator: progress and bottlenecks%
\subsection{LLM-driven Text Quality Evaluators}

To advance fine-grained evaluation in NLG, researchers have developed both specialized and unified assessment frameworks. Dimension-specific evaluators now target critical quality aspects such as summarization consistency \cite{kryscinski2021evaluating,wang2020asking} and dialogue coherence \cite{dziri2019evaluating,huang2020grade}, while unified approaches employ diverse methodologies including adaptive prompting \cite{yuan2021bartscore}, model ensembles \cite{mehri2020usr}, and dynamic metric integration \cite{scialom2021questeval} to generate comprehensive quality assessments.

While these methods demonstrate versatility, their lack of explainability remains a significant limitation, often reducing complex evaluations to simplistic summative metrics \cite{li2025unieval}. In response, the field has witnessed three key developments in transparent evaluation architectures: 1) rationale-enhanced systems like G-EVAL \cite{liu2023g} and UNIEVAL \cite{li2025unieval} that combine scoring with interpretable reasoning chains; 2) reference-free paradigms including GPTScore's probability-based quantification \cite{fu2023gptscore} and PandaLM's efficient comparative assessment \cite{wang2023pandalm}; and 3) robust evaluation frameworks such as JudgeLM \cite{zhu2023judgelm} designed for adversarial testing scenarios. These innovations, while significantly improving evaluation transparency, have inadvertently created new challenges regarding computational efficiency and scoring flexibility.

The current evaluation landscape exhibits two major limitations. First, the reliance on large language model–based architectures entails prohibitive computational demands \cite{liu2023g,fu2023gptscore}. Second, existing scoring mechanisms often reduce complex distinctions to overly simplistic aggregates or rely on rigid weighting schemes \cite{li2025unieval}. Although general-purpose evaluators have advanced considerably, systematic approaches to domain-specific quality assessment, particularly in legal contexts, remain insufficiently developed.

\subsection{Text Quality Evaluation in The Legal Domain}

Despite the growing interest in general-purpose quality evaluators, the assessment of LLM-generated legal texts remains relatively underexplored. Initial research in this domain primarily concentrated on evaluating the legal reasoning capabilities of language models. For instance, Guha et al.~\cite{guha2023legalbench} introduced LegalBench, a comprehensive benchmark targeting tasks related to legal memory, understanding, and reasoning. Subsequent efforts, including LawBench~\cite{fei2023lawbench} and LAiW~\cite{dai2024laiwchineselegallarge}, extended this evaluation paradigm to a broader range of jurisdictions and legal subdomains.

Beyond LLM-focused benchmarks, earlier interdisciplinary research at the intersection of linguistics and law has long examined the quality of legal texts. As early as the 1990s, scholars explored linguistic and stylistic dimensions of legal drafting and statutory clarity~\cite{fajans1998linguistics, levi1995introduction}. Parallel movements such as the “plain English” advocacy emphasized accessibility and readability in legal communication, exemplified by Kimble’s work on plain language in law~\cite{kimble1996writing}. Complementing this, Fry analyzed the legal aspects of readability and its implications for legal drafting~\cite{fry1998legal}. Later, Waltl and Matthes systematically analyzed the structural and linguistic complexity of German laws, introducing quantitative measures of legal text difficulty~\cite{waltl2014towards}. More recently, Han et al. conducted a systematic literature review of readability metrics in legal texts, highlighting how quantitative indices can capture issues of clarity, readability, and logical structure in statutory documents~\cite{han2024use}. These works demonstrate that legal text quality can be evaluated along dimensions of complexity, clarity, and linguistic coherence, even though they focus on human-authored texts rather than LLM-generated content. In contrast, our framework specifically addresses the quality evaluation of LLM outputs, thereby bridging this gap.

To measure legal accuracy, researchers have employed both classical text similarity metrics (e.g., BLEU, ROUGE)~\cite{wang2024legalevalutionschallengeslarge} and legal-specific benchmarks such as LexGLUE~\cite{chalkidis2022lexgluebenchmarkdatasetlegal}, which provides multiple tasks for assessing legal language understanding. In parallel, JuDGE~\cite{chlapanis2025greekbarbenchchallengingbenchmarkfreetext} introduced a challenging free-text benchmark simulating bar exam-style questions to further stress-test LLM legal competencies.

More recent work has shifted attention toward evaluating model trustworthiness and fairness. LexEval~\cite{li2024lexevalcomprehensivechineselegal} and FairLex~\cite{chalkidis2022fairlex} represent emerging efforts to audit the reliability and equity of legal model outputs, particularly in multilingual and multi-jurisdictional contexts.

However, despite these developments, comprehensive evaluation frameworks targeting the textual quality of LLM-generated legal content, particularly in terms of logical consistency and structural completeness, remain scarce. To address this gap, we propose a decoupled evaluation architecture that integrates a fine-tuned legal quality assessor with a regression-based scoring module. This approach facilitates efficient, explainable, and granular assessments of legal text quality, offering a scalable alternative to current manually intensive evaluation pipelines.

\section{Proposed Method}

This work proposes a domain-specific evaluation framework for assessing the quality of LLM-generated legal texts. Unlike general-purpose evaluators that focus on fluency or coherence, our method targets key legal dimensions such as logical consistency, structural completeness, and linguistic expression. In our framework, logical consistency is subsumed under Content Quality, while structural completeness is reflected in Structure and Organization, ensuring alignment between the overarching legal requirements and the five quality dimensions introduced below.

Legal texts exhibit distinctive characteristics, including rigid structural conventions, highly specialized terminology, and stringent logical requirements. These features necessitate evaluation criteria that are specifically adapted to the legal domain. The proposed framework is designed to meet these needs by ensuring contextual relevance and providing interpretable feedback. Although the present implementation focuses on Chinese legal texts, both the comment generation and scoring components are language-agnostic and can be extended to other jurisdictions through corpus substitution and adjustments in terminology. 

\subsection{Quality Criteria}

\begin{figure}[htbp]
  \centering
  \includegraphics[width=0.9\textwidth]{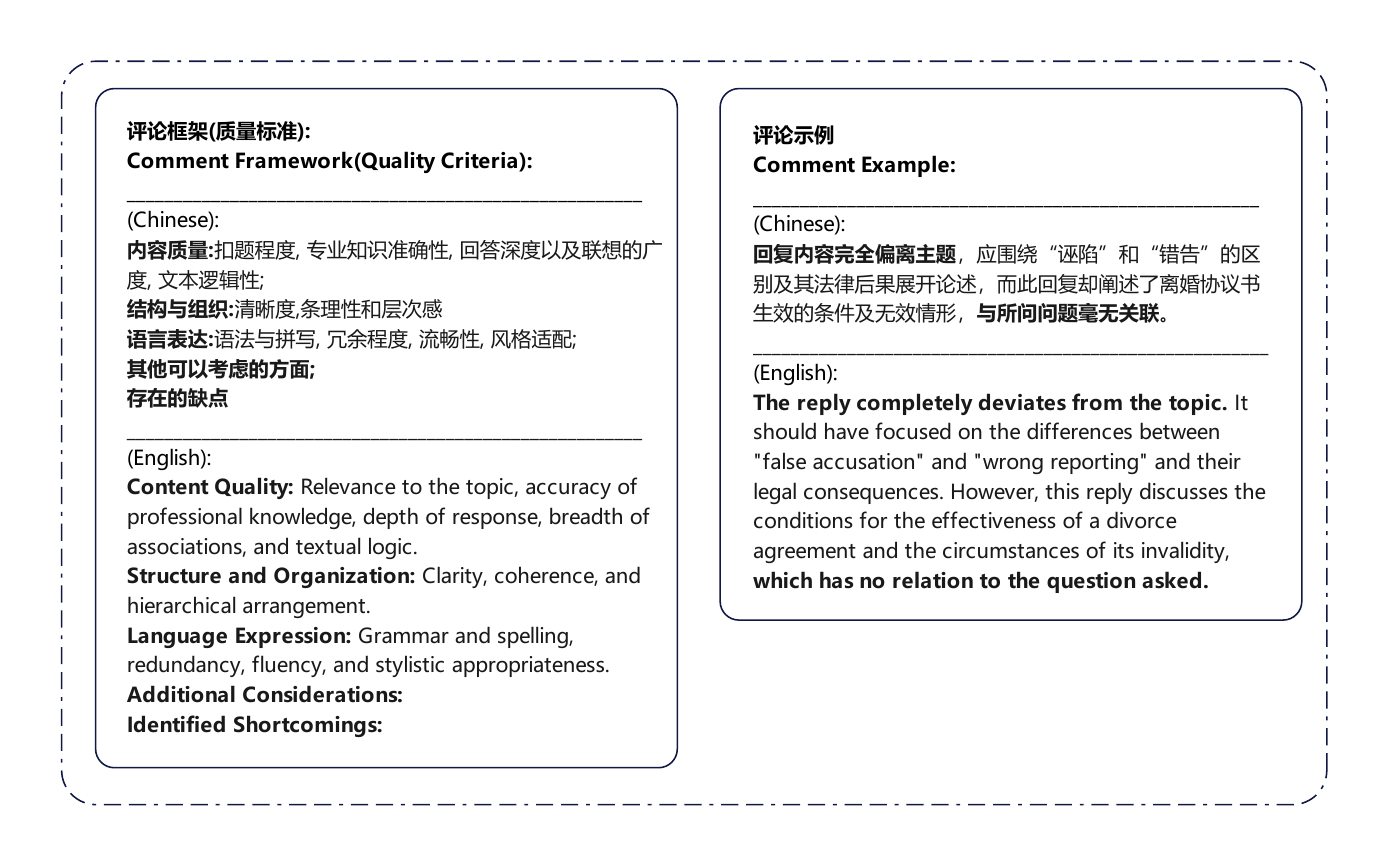}
  \caption{Illustration of how evaluation criteria are mapped into comments. 
    The figure serves as a demonstration of the mapping process rather than a complete input–output case. 
    The example is presented in the original Chinese text together with its English translation.
    Full examples with complete input–output pairs are provided in the appendix B.}
  \label{fig:figure_1_Comment_framework_and_examples}
\end{figure}

A central component of our framework is the definition of five quality dimensions that provide a conceptual foundation for analyzing the linguistic performance of legal texts. These dimensions are not designed as rigid rubrics with fixed weights or scoring tables. Instead, they serve as reference axes, encouraging evaluators to consider multiple perspectives while retaining the flexibility of subjective judgment. This design allows us to capture authentic reading habits and commentary styles, reflecting how ordinary readers may value conciseness, comprehensiveness, or stylistic naturalness in different ways.

\textbf{Content Quality:} Relevance to the topic, accuracy of professional knowledge, depth of response, breadth of associations, and textual logic. This dimension emphasizes whether an answer is both legally correct and sufficiently comprehensive.  

\textbf{Structure and Organization:} Clarity, coherence, and hierarchical arrangement. This dimension highlights whether reasoning is presented in a well-structured and logically ordered manner.  

\textbf{Language Expression:} Grammar and spelling, redundancy, fluency, and stylistic appropriateness. This dimension focuses on the linguistic surface form and its suitability for professional legal contexts.  

\textbf{Additional Considerations:} Other context-sensitive aspects that may affect evaluation, such as the explicitness of assumptions, citation style, or clarity of limitations.  

\textbf{Identified Shortcomings:} Explicit documentation of observed weaknesses across the above dimensions to support systematic error analysis.  

Figure~\ref{fig:figure_1_Comment_framework_and_examples} illustrates how these criteria are mapped into annotator comments through a simple example. It should be noted that this figure serves as a \emph{demonstration of the mapping process} rather than a complete input–output case. Full examples, including query, answer, comment, conclusion, and score, are provided in Appendices B (covering representative cases at scores 0, 58, and 96).

\subsection{Evaluation Data Design}

The dataset serves as the foundation for training and validating our evaluation framework, as well as for systematically benchmarking diverse LLMs. To maximize evaluative reliability while controlling annotation costs, we design a small-scale yet high-quality dataset, structured into five interdependent components: query, answer, comment, conclusion, and score.

To ensure diversity and robustness, the training data was generated using models spanning parameter sizes from 26M to 685B, covering a wide range of architectures (Table~\ref{tab:models}). In addition, we included specially constructed Error-Input cases, where queries are deliberately paired with irrelevant inputs to generate mismatched answers. These serve as negative samples (score = 0), ensuring that the evaluation model learns to penalize responses that, while potentially fluent, fail to address the given query.

\begin{table}[htbp]
 \centering
 \caption{Models for Training Data Generation and Their Parameter Sizes.}
 \label{tab:models}
 \begin{tabular}{lc}
  \toprule
  \textbf{Model Name} & \textbf{Parameter Size} \\
  \midrule
  MiniMind2-small~\cite{Jing2025MiniMind} & 26M \\
  Qwen2.5-0.5B-GPTQ-Int4~\cite{qwen2025qwen25technicalreport} & 0.49B (Int4) \\
  Qwen2.5-0.5B~\cite{qwen2025qwen25technicalreport} & 0.49B \\
  Qwen2.5-1.5B~\cite{qwen2025qwen25technicalreport} & 1.5B \\
  Qwen2.5-3B~\cite{qwen2025qwen25technicalreport} & 3.1B \\
  Qwen2.5-14B~\cite{qwen2025qwen25technicalreport} & 14.7B \\
  Doubao-1.5-Pro-32k~\cite{Bytedance2025doubao15pro} & Unpublished\textsuperscript{*} \\
  Deepseek-R1~\cite{deepseekai2025deepseekv3technicalreport} & 671B \\
  Deepseek-V3-0324~\cite{deepseekai2025deepseekv3technicalreport} & 685B \\
  Error-Input & None\textsuperscript{**} \\
  \bottomrule
  \multicolumn{2}{l}{\footnotesize 
   \begin{tabular}{@{}p{\linewidth}@{}}
    \textsuperscript{*} \textit{Unpublished}: Official parameter count not disclosed. Estimated active parameters: 20B (MoE). \\
    \textsuperscript{**} \textit{None}: Randomly select a random answer from another query and fill it in to simulate the situation where the answer is irrelevant or the relevance of the answer is not high. \\ 
    \textsuperscript{**} \textit{Error-input/None}: Randomly select a random answer from another query and fill it in to simulate the situation where the answer is irrelevant or the relevance of the answer is not high. \\ 
   \end{tabular}
  }
 \end{tabular}
\end{table}

\textbf{Query:}
In this study, a "query" refers to the input legal issue presented to the evaluated model, with all queries drawn from Chinese legal domains including criminal law, civil code, and general statutes. We initially sampled 10,000 queries from an integrated collection comprising the DISC-Law-SFT-Pair dataset \cite{yue2024lawllm} for general legal matters, the Criminal-Law-Dataset \cite{yuan2024Crimina} for criminal cases, and our proprietary dataset focused on the Civil Code of the People's Republic of China. This large pool provided sufficient coverage and diversity, from which the annotated dataset of 946 queries was later derived.

\textbf{Answer:} 
An "answer" refers to the output produced by an evaluated model in response to a given query. Unlike the training dataset, where answers are pre-generated by selected models, the evaluation dataset does not contain pre-defined responses. Instead, all answers are generated in real time by the evaluated models during testing. For completeness, special cases such as Doubao-1.5-Pro-32k and Error-Input are also included to ensure architectural diversity and to introduce noise control.

\begin{figure}[htb]
  \centering
  \includegraphics[width=0.9\textwidth]{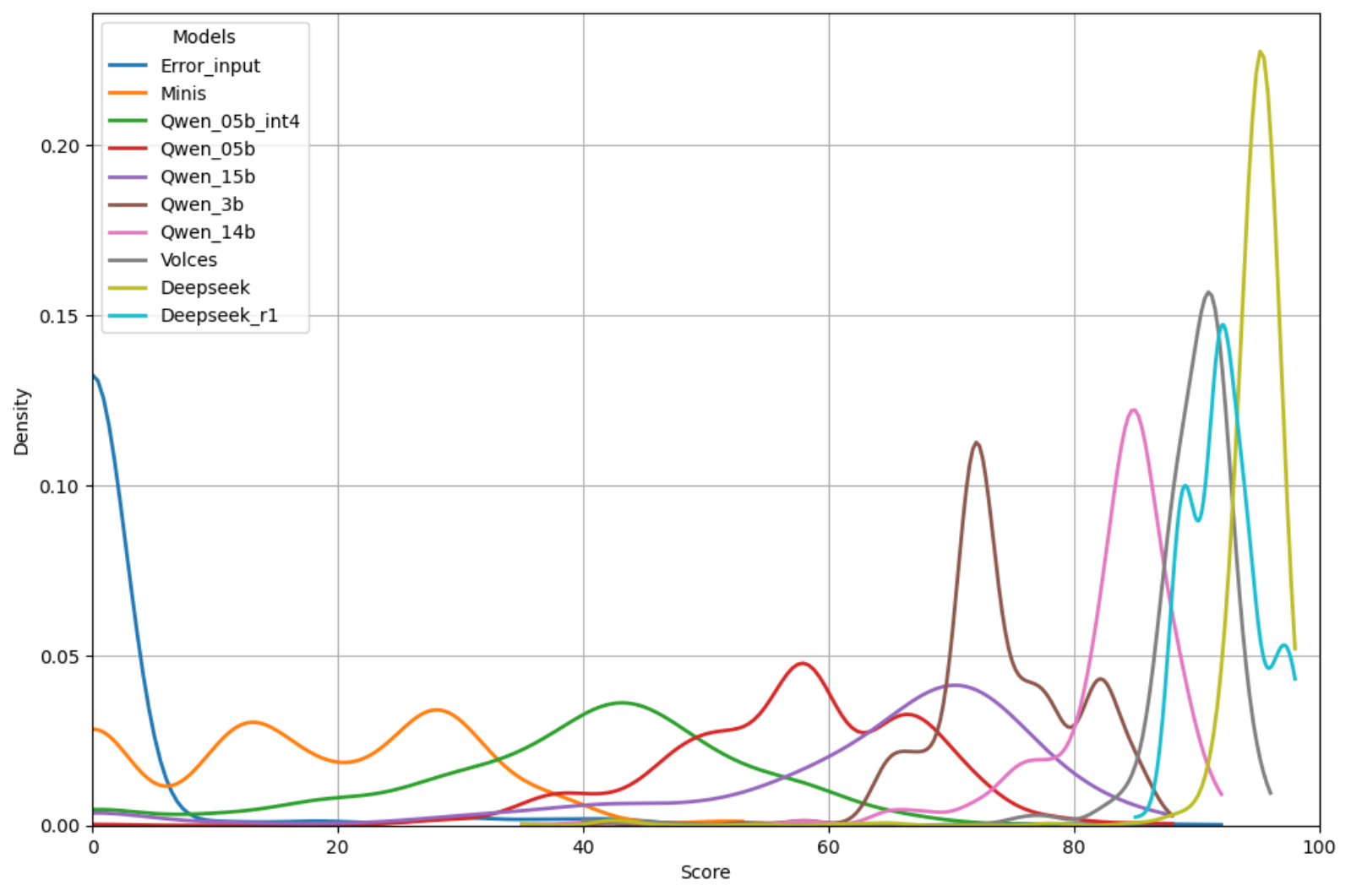}
  \caption{Distribution of model scores with probability density, illustrating central tendency and variance in legal text quality evaluation.}
  \label{fig:probability_score}
\end{figure}

\textbf{Comment:}
A "comment" represents a meta-evaluation of answer quality, manifesting as textual feedback that critically analyzes response content. As illustrated in Figure~\ref{fig:figure_1_Comment_framework_and_examples}, the comment framework systematically evaluates answers through five key dimensions: 1) \textit{content quality} (including topical relevance and professional accuracy); 2) \textit{structure and organization}; 3) \textit{linguistic expression}; 4) \textit{additional considerations}; 5) \textit{identified deficiencies}. Our evaluation framework builds upon foundational legal text quality frameworks \cite{davoodijam2024evaluation}, while specifically adapting the assessment dimensions to address unique legal domain requirements.

These comments serve two closely connected functions: (1) they act as qualitative assessments that identify deficiencies in answers through domain-specific critique and (2) they provide the foundational training corpus for developing the comment model's reasoning module. While adhering to the evaluative framework that ensures methodological consistency, the comments also retain sufficient linguistic diversity to cover the full range of possible answer quality issues, thereby enabling the model to acquire refined judgment capabilities.

\textbf{Conclusion:} 
A "conclusion" in our dataset synthesizes the comprehensive evaluation of an answer into a summative statement that provides global quality assessment. This component serves three critical functions: 1) it consolidates the multi-dimensional analysis from the comment into an overarching judgment (such as "While demonstrating professional terminology usage, the answer substantially deviates from the core legal issues"); 2) it trains the evaluation model's scoring module by supplying holistic quality references rather than direct score mappings; and 3) it completes the evaluation framework's logical flow from query analysis to final assessment. The conclusion's global perspective ensures the model considers all quality dimensions simultaneously when generating scores, while its non-prescriptive nature maintains flexibility in the final scoring decision.

\textbf{Score:} 
The "Score" serves as the final quantitative output of this study, measuring the legal text quality level of a model's responses. It is derived from three key inputs: 1) the original answer text, which provides direct semantic and structural details; 2) the Comment, offering a fine-grained analysis of strengths and weaknesses; and 3) the Conclusion, synthesizing these observations into a holistic assessment. We calculate scores on a 0–100\% scale, where higher values indicate better text quality and lower values indicate poorer quality, ensuring a standardized and interpretable metric. Figure~\ref{fig:probability_score} illustrates the score distributions across all models in the training set, along with trends in average performance and variance.

Unlike prior approaches that directly collapse text evaluation into a single summative number, our score is not intended as a simplistic stand-alone metric. Instead, it functions as the third layer of our evaluation framework, building on the explanatory foundation of Comments and Conclusions. The Comments provide fine-grained diagnostic insights, the Conclusion synthesizes them into an accessible qualitative judgment, and the Score then calibrates this judgment into a standardized quantitative form. This layered design ensures that the numerical score remains interpretable and tightly connected to qualitative evidence, while still enabling systematic benchmarking and cross-model comparison. In this way, our framework explicitly addresses the limitation highlighted in prior work, namely that complex evaluations are often reduced to opaque single metrics, by ensuring that quantitative results are always grounded in transparent qualitative analysis.

\subsection{Evaluation Framework: Model Architecture and Process}

% 在第一个段落结束后插入
\begin{figure*}[htbp]
  \centering
  \includegraphics[width=0.9\textwidth]{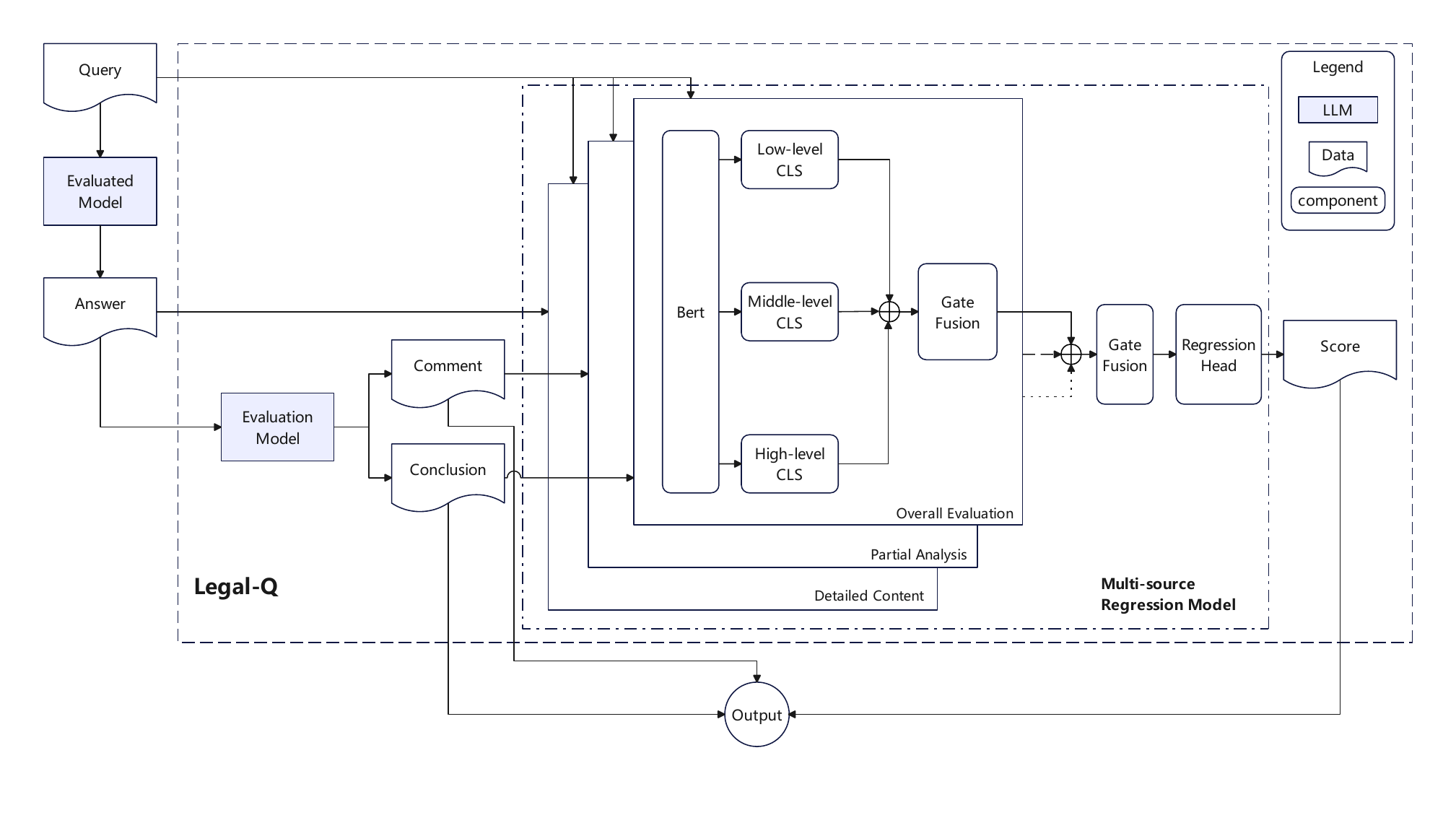}
  \caption{End-to-end data flow of the evaluation framework, showing how raw inputs are processed through comment generation, conclusion synthesis, and final scoring modules. 
  Abbreviation: Classification Token (CLS), representing sequence-level embeddings extracted from transformer layers.}
  \label{fig:figure_3_model_constructig}
\end{figure*}

The Legal-Q framework does not generate answers itself; rather, it systematically evaluates outputs from external models. In this study, 49 candidate LLMs serve as input sources, while Legal-Q provides the assessment pipeline shown in Figure~\ref{fig:figure_3_model_constructig}. The process begins with external evaluated models, which in this study correspond to 49 candidate large language models used to generate answers to legal queries. These evaluated models are not part of the Legal-Q system itself but provide the inputs that the framework analyzes.

The first component of Legal-Q is the Evaluation Model, which is implemented as a fine-tuned Qwen2.5-7B. This model processes the answers generated by the evaluated models and produces structured Comments and Conclusions. To enhance interpretability and ensure logically consistent judgments, the Evaluation Model employs chain-of-thought (CoT) reasoning. This approach explicitly generates intermediate reasoning steps before finalizing its outputs~\cite{nye2021show, wei2022chain}. By explicitly modeling reasoning trajectories, the system is able to articulate why certain strengths or weaknesses are identified. The Comments thus provide Partial Analysis by offering qualitative and dimension-specific feedback that critically examine relevance, accuracy, structure, and expression. The Conclusions then provide an Overall Evaluation by synthesizing these observations into a holistic judgment that captures the overall quality of the response. Together, these two outputs represent the qualitative aspect of the framework.

The second component, termed the Multi-source Regression Model, is statistically equivalent to a logistic regression classifier~\cite{ng2001discriminative} operating on embeddings from multiple feature streams. It integrates heterogeneous inputs from three complementary sources: the Answer as Detailed Content, the Comment as Partial Analysis, and the Conclusion as Overall Evaluation. Each input is first encoded by a dedicated BERT model~\cite{vaswani2017attention}, from which we extract the Classification Token(CLS) representations, widely adopted as compact sequence-level embeddings that serve as proxies for the global semantics of text sequences~\cite{reimers2019sentencebertsentenceembeddingsusing}. By leveraging hierarchical CLS extraction across multiple layers, the model captures both shallow lexical cues and deeper semantic features. Logistic regression is adopted here because it provides a transparent and statistically grounded way to combine multiple feature streams into a single prediction~\cite{lavalley2008logistic}. Concretely, the embeddings are linearly combined and then passed through a Sigmoid activation, which maps the unbounded linear score into a normalized range of [0,1]~\cite{ayyadevara2018logistic}. This probability-like output is subsequently rescaled to (0, 100), enabling its interpretation as a percentage-style quality score.

To further enhance feature integration, we employ gated fusion mechanisms, which dynamically adjust the relative contributions of Answer, Comment, and Conclusion through learnable gates. Unlike simple concatenation, gated fusion selectively emphasizes informative signals while suppressing noise, thereby yielding a more stable regression outcome~\cite{li2020gated}. This design highlights that while we use the shorthand “Multi-source Regression Model,” its underlying structure is equivalent to a logistic regression model operating on multiple feature streams. For clarity, we emphasize that the outputs of Legal-Q are not limited to a single quantitative score. Instead, the system consistently generates three outputs: a Comment, a Conclusion, and a Score, ensuring both qualitative interpretability and quantitative rigor. This multi-output design contrasts with prior evaluators that collapse quality into a single opaque score, thereby ensuring both diagnostic depth and benchmarking utility.

\subsection{Data Annotation and Training}

All components of the evaluation framework, namely Comments, Conclusions, and Scores, were developed through a structured annotation pipeline that combined AI-assisted drafting with staged human review. A small group of legal advisors, including practicing professionals and legal research staff, contributed to the definition of evaluation standards, the preparation of the annotation handbook, and the validation of pilot cases. The large-scale annotation was conducted by four trained university students under continuous supervision. Annotators received systematic preparation that included trial batches, targeted feedback from advisors, and embedded quality-control items. Reliability was verified through repeated cases and error-injected items, while inter-annotator agreement was periodically measured to ensure consistency.

From the initial pool of 10,000 sampled queries, we constructed an annotated dataset comprising 946 queries. Empirically, we observed that model performance and annotation stability converged with this subset size, making further expansion unnecessary. Each annotated query was associated with multiple model-generated responses and error-injected baselines, yielding approximately 9,460 annotated items in total.

The annotated data served as the foundation for training the Legal-Q framework through a two-stage process. In the first stage, the Evaluation Model was fine-tuned on (Query, Comment, Conclusion) instances, which enabled Qwen2.5-7B to learn to generate evaluative feedback that was consistent with the defined annotation framework. In the second stage, the Multi-source Regression Model was trained using complete (Query, Answer, Comment, Conclusion, Score) tuples. The regression component was optimized with multi-task objectives that combined score prediction with auxiliary feature reconstruction. This strategy ensured methodological alignment between qualitative and quantitative components, while also enhancing robustness and preserving computational efficiency. Together, these training steps allowed Legal-Q to provide reliable, interpretable, and reproducible assessments of legal text quality in the legal domain.

\subsection{Stability-Adjusted Evaluation Metric}

To jointly evaluate model capability and prediction reliability, we propose a novel composite metric, AdjScore, which combines the mean quality score and its variability based on the Coefficient of Variation (CV)\cite{brown1998coefficient}. Inspired by risk-adjusted measures such as the Sharpe Ratio\cite{sharpe1994sharpe}, AdjScore employs a distinct nonlinear formulation to balance performance and stability. Consider a model $f$ evaluated on $N$ test samples ${s_i}_{i=1}^N$:

\begin{equation}
\mu = \frac{1}{N}\sum_{i=1}^N s_i
\end{equation}

\begin{equation}
\sigma = \sqrt{\frac{1}{N}\sum_{i=1}^N (s_i - \mu)^2}
\end{equation}

The Coefficient of Variation (CV) normalizes dispersion relative to mean performance:

\begin{equation}
\mathrm{CV} = \sigma/\mu \quad \in [0, +\infty)
\end{equation}

Our stability-adjusted score then combines these components:

\begin{equation}
\mathrm{AdjScore} = \frac{\mu}{1 + \mathrm{CV}} = \frac{\mu^2}{\mu + \sigma} \quad \in (0, \mu)
\end{equation}

The stability-adjusted scoring metric provides a principled approach to model evaluation by jointly quantifying average quality and prediction reliability. By incorporating the coefficient of variation, the metric balances mean performance with output consistency, ensuring that models achieving high scores demonstrate not only strong average performance but also stable behavior across diverse inputs. This property is particularly important in legal text generation, where consistency in tone, terminology, and reasoning quality underpins the reliability and predictability of professional practice. The non-linear formulation naturally penalizes high-variance predictions that may indicate unstable behavior, while the bounded range $(0,\mu)$ guarantees that AdjScore never exceeds the mean score, providing a conservative and interpretable measure. In addition, the decomposition into $\mu$ and $\sigma$ components offers diagnostic insight by distinguishing whether poor performance arises from inherent capability limitations or from undesirable output fluctuations. For practical deployment, the coefficient of variation functions as an effective reliability filter, enabling practitioners to identify models that maintain both high quality and stability. This makes AdjScore particularly suitable for ranking candidate models and guiding cost–performance trade-offs in applied settings.

\section{Experiments} \label{sec:experiments}

In the experimental part, we used a validation set with the same data source distribution as the training set but strictly non-overlapping content. In order to evaluate the legal text quality performance of more models more quickly, the number of validation sets content is only 60. All experiments were conducted in a Python 3.12.9 (Anaconda) environment with PyTorch 2.6.0 and CUDA 12.4, utilizing 8xNVIDIA GeForce RTX 4090 GPUs for acceleration. Model inferences were performed using default generation parameters (temperature, top-p, and max new tokens) unless otherwise specified. 

In total, we systematically evaluated 49 models spanning parameter scales from 26M to 685B across diverse architectures. For readability, the main text presents representative subsets of results, while the complete ranking and detailed statistics for all 49 models are reported in Appendix~A.

\subsection{Results Analysis}

Before presenting the numerical results, it is important to clarify the role of the score in our analysis. The score should not be viewed as an isolated or reductive number; rather, it represents the calibrated quantitative layer of a three-part evaluation pipeline. Its purpose is twofold: (i) to provide a standardized benchmark for cross-model ranking and large-scale comparison, and (ii) to remain grounded in the qualitative evidence offered by Comments and Conclusions. In this way, our framework avoids the pitfall of reducing complex evaluations to opaque single metrics, while still enabling reproducible and comparable assessments.

\begin{table}[htbp]
 \centering
 \caption{Cross-Set Performance Evaluation of Language Models using Score (Mean). 
 $\Delta$ denotes the difference between validation and training scores (Validation – Training).}
 \label{tab:experiements}
 \begin{tabular}{lrrr}
  \toprule
  \textbf{Model} & \textbf{Training (Mean)} & \textbf{Validation (Mean)} & \boldmath{$\Delta$} \\
  \midrule
  MiniMind2-small & 17.66 & 23.23 & +5.57 \\
  Qwen2.5-0.5B-GPTQ-Int4 & 38.92 & 45.81 & +6.89 \\
  Qwen2.5-0.5B & 56.93 & 55.31 & -1.62 \\
  Qwen2.5-1.5B & 61.76 & 68.08 & +6.32 \\
  Qwen2.5-3B & 74.39 & 76.83 & +2.44 \\
  Qwen2.5-14B & 82.89 & 80.05 & -2.84 \\
  Doubao-1.5-Pro-32k & 89.41 & 89.59 & +0.18 \\
  Deepseek-R1 & 92.32 & 93.76 & +1.44 \\
  Deepseek-V3-0324 & 94.36 & 94.09 & -0.27 \\
  \bottomrule
 \end{tabular}
\end{table}

Table~\ref{tab:experiements} reports results based on Score (Mean), which represents the average quality score across all evaluation items for each model. A clear dichotomy emerges by scale. For low-scoring models below 70 points (MiniMind2-small and Qwen2.5-0.5B variants), we observe substantial cross-set fluctuations, with MiniMind2-small showing a fluctuation of +5.57 points and Qwen2.5-0.5B-GPTQ-Int4 of +6.89 points. These models exhibit standard deviations exceeding 7 points (versus less than 3 points for high-scoring models), reflecting both imbalanced training data distribution and inherent architectural limitations.

In contrast, larger models display remarkable stability, exemplified by Qwen2.5-14B’s modest decline of -2.84 points and the top-performing Deepseek-V3-0324’s minimal change of -0.27 points. Overall, the high-score group maintains an average cross-set error of just 1.2 points (compared to 12.7 points for the low-score group), confirming a strong correlation between model scale and evaluation stability. 

In addition to the above stability analysis, we also performed a cross-linguistic validation experiment to examine whether the framework maintains consistent behavior when applied to Chinese and English. Specifically, as detailed in Appendix C, we translated a subset of Chinese queries and model-generated answers into English and re-evaluated them under identical conditions. The paired $t$-test yielded $p = 0.3781$ with a negligible effect size of $-0.038$, and the Wilcoxon signed-rank test produced consistent non-significant results. A mixed-effects model controlling for item-level variance likewise showed that the fixed effect of language (English) was not significant ($\beta = 0.286$, $p = 0.846$). Taken together, these results indicate that the framework provides consistent assessments in both Chinese and English, offering preliminary evidence for its cross-linguistic applicability.

While these results based on Score (Mean) reveal broad performance patterns across model scales, they do not fully capture the role of prediction stability. In particular, models with similar mean scores may behave quite differently in terms of variance. To address this limitation, we next evaluate models using the stability-adjusted metric, AdjScore, which integrates both mean performance and consistency into a single measure.

\subsection{Leaderboard}

Building on the Score (Mean) analysis above, Table~\ref{tab:benchmark} reports results using the stability-adjusted metric, \textbf{AdjScore}. For ease of discussion, we group models into three rough tiers based on their observed AdjScore ranges ($\geq 90$, 70--90, $<70$)
. These tiers are descriptive rather than formal classifications and serve as convenient labels for highlighting performance and stability differences. In particular, AdjScore reveals finer-grained contrasts among models with similar mean scores, distinguishing stability trade-offs in the mid tier and error modes in the low tier.

\begin{table}[h]
 \centering
 \caption{Model Performance Leaderboard based on Score (Mean), standard deviation (Std), and stability-adjusted score (AdjScore).
 For clarity of discussion, models are loosely grouped into three tiers (High, Mid, Low) according to their observed AdjScore ranges, without implying a formal classification standard.}
 \label{tab:benchmark}
 \begin{tabular}{lccc}
  \toprule
  \textbf{Model} & \textbf{Score (Mean)} & \textbf{Std} & \textbf{AdjScore} \\
  \midrule
  \addlinespace
  \multicolumn{4}{@{}l}{\textit{High Tier (AdjScore $\geq 90$)}} \\
  Deepseek-v3-0324 & 94.09 & 0.93 & 93.16 \\
  Qwen-QWQ-32B & 94.06 & 1.06 & 93.01 \\
  Deepseek-R1 & 93.76 & 1.23 & 92.54 \\
  Qwen3-14B & 93.25 & 1.17 & 92.09 \\
  \addlinespace
  \multicolumn{4}{@{}l}{\textit{Mid Tier (AdjScore 70–90)}} \\
  ERNIE\_4.5 & 89.07 & 4.99 & 84.34 \\
  Volces-Pro & 89.59 & 5.73 & 84.21 \\
  Deepseek-v3(old) & 87.27 & 7.08 & 80.72 \\
  qwen2.5\_32B & 80.18 & 8.05 & 72.87 \\
  Qwen2.5-7B & 79.60 & 7.60 & 72.66 \\
  Kimi-128k & 75.67 & 6.94 & 69.32 \\
  ChatGPT-4o & 74.56 & 7.75 & 67.54 \\
  \addlinespace
  \multicolumn{4}{@{}l}{\textit{Low Tier (AdjScore $<$ 70)}} \\
  ChatGPT-3.5 & 66.11 & 7.74 & 59.18 \\
  Qwen2.5-1.5B & 68.08 & 11.15 & 58.49 \\
  Qwen2.5-0.5B & 55.31 & 12.43 & 45.16 \\
  MiniMind2 & 23.23 & 12.55 & 15.08 \\
  Qwen2.5-Math-15B & 13.58 & 12.99 & 6.94 \\
  \bottomrule
 \end{tabular}
\end{table}

The top-tier models demonstrate exceptional consistency: Deepseek-v3-0324 achieves the highest AdjScore (93.16) by combining near-perfect mean performance (94.09) with minimal fluctuation (Std=0.93). Comparable mean-scoring models like Qwen-QWQ-32B (94.06) show slightly reduced AdjScore (93.01) due to marginally higher variance (Std=1.06). 

Notably, the metric effectively captures stability trade-offs in mid-tier models. While ERNIE\_4.5 exhibits lower mean performance (89.07) than Volces-Pro (89.59), its superior stability (Std=4.99 vs 5.73) results in a higher AdjScore (84.34 vs 84.21). This aligns with our metric's design principle in Eq.~(4), where the non-linear $\mu^2/(\mu+\sigma)$ formulation penalizes high-variance predictions.

The bottom-tier models reveal two characteristic error modes: (i) consistently low performance (Qwen2.5-Math-15B: Mean=13.58) and (ii) unstable mid-range performers (Qwen2.5-1.5B: Mean=68.08 but Std=11.15). This bifurcation suggests that AdjScore provides distinct signals for both inherent capability limitations and reliability concerns. Appendix~A reports the full leaderboard with all 49 models for reference.

\section{Discussions}

The preceding experiments establish the comparative performance of 49 models under our evaluation framework. In this section, we move beyond raw benchmarking results to analyze the factors that drive model behavior and to derive practical insights for deployment. We divide the discussion into two complementary perspectives: intrinsic performance factors that reflect model-internal properties, and engineering trade-offs that capture external constraints and deployment considerations.

\subsection{Model Performance Factors}

This subsection investigates intrinsic properties of large language models that affect their performance in legal text evaluation. We focus on four dimensions: (i) model scale, (ii) quantization, (iii) context length, and (iv) model type (reasoning vs. base). Each factor is examined independently to clarify its specific contribution to overall quality and stability.

\subsubsection{Effect of Model Scale}

\begin{figure}[h]
  \centering
  \includegraphics[width=0.9\textwidth]{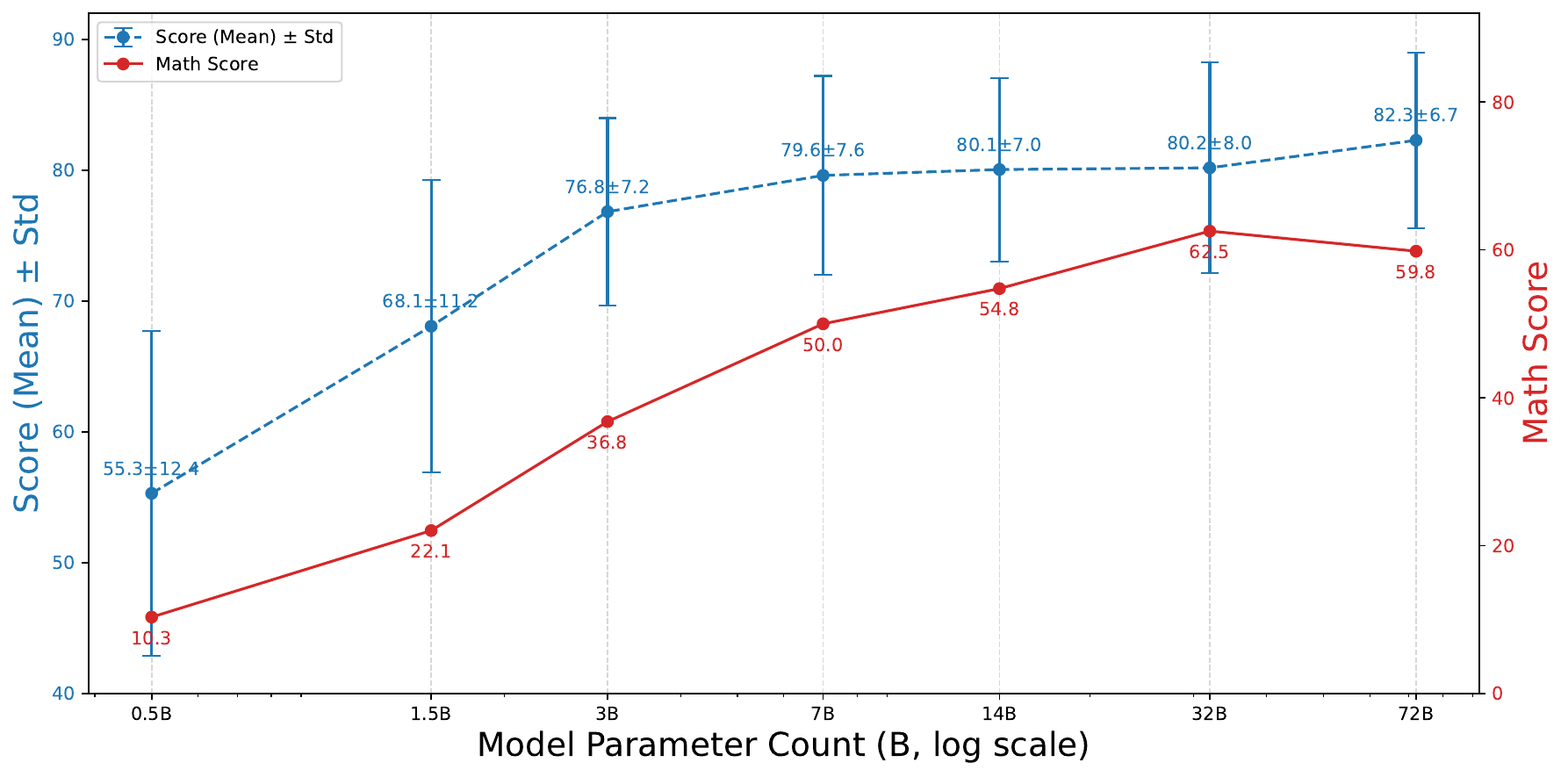}
  \caption{Scaling laws of model capabilities: legal text quality (blue) vs. mathematical reasoning (red) across 0.5B–72B models.}
  \label{fig:fig_dis_f2_Score}
\end{figure}

Systematic evaluation across seven model scales (0.5B to 72B parameters) reveals a logarithmic relationship between parameter count and capability metrics (Figure~\ref{fig:fig_dis_f2_Score}). The 72B variant shows only a marginal 2.7\% improvement over its 7B counterpart (p=0.032, paired t-test), with overlapping 95\% confidence intervals confirming performance saturation beyond 7B parameters. Importantly, the scaling behavior is not uniform across tasks: legal text quality plateaus at 7B parameters, whereas reasoning capability continues to improve until 14B, as illustrated by the MATH benchmark (12,500 competition-level problems with step-by-step solutions~\cite{Hendrycks2021Measuring}). These results suggest that model size exerts a stronger impact on reasoning than on expressive quality. These results suggest that model size exerts a stronger impact on reasoning than on expressive quality. Notably, within the Qwen3 family, the 4B variant does not lie on the trade-off frontier, as its performance is slightly lower than that of the smaller 1.7B model.

The observed nonlinear scaling patterns warrant further examination. Together, the MATH benchmark, which quantifies intrinsic reasoning, and the legal text quality metric, which evaluates expressive fidelity, reveal diminishing returns in the Qwen2.5 series models at an early stage. We attribute this phenomenon to two constraints: the linguistic quality ceiling of the training data, which limits expressive potential, and architectural bottlenecks, which restrict reasoning scalability. Overall, these findings indicate that brute-force scaling beyond 7B offers limited returns under cost constraints, emphasizing the importance of architectural efficiency and data quality improvements.

\subsubsection{Effect of Quantization}

Model quantization is a widely used technique in the deployment of large-scale neural networks. The central idea is to reduce the numerical precision of weights and activations from high-precision floating-point formats (e.g., FP32 or BF16) to lower-precision formats such as Int8 or Int4. This reduction significantly decreases memory consumption and improves inference speed, while often maintaining comparable model accuracy within acceptable bounds~\cite{egashira2024exploiting}. Quantization has therefore become an indispensable tool for practical deployment of LLMs, especially in scenarios where hardware resources and energy efficiency are critical~\cite{hasan2024optimizing}. In this subsection, we specifically examine how quantization affects the legal text quality of LLMs.

In this study, we use Qwen2.5-7B-Instruct as a case model to investigate the impact of quantization on legal text quality. We consider three precision formats:

\textbf{BF16} (Brain Floating Point 16,in bfloat16): A 16-bit floating-point format designed to balance dynamic range and numerical efficiency. It is widely supported on modern accelerators and often serves as a baseline reduced-precision representation, which is the  the original models  precision of Qwen2.5 series~\cite{qwen2025qwen25technicalreport}.

\textbf{Int8} (8-bit integer quantization): A commonly used quantization scheme that compresses parameters and activations to 8-bit integers. Int8 typically yields substantial efficiency gains while maintaining accuracy close to BF16~\cite{dettmers2022gpt3}.

\textbf{Int4} (4-bit integer quantization): A more aggressive quantization method that reduces memory and computation further. However, due to its extremely limited representational capacity, Int4 may introduce noticeable performance degradation, particularly in tasks requiring fine-grained linguistic precision~\cite{Xiaoxia2023Understanding}.

Table~\ref{tab:Quantization_1} reports performance across four benchmarks. In addition to the experimental data of this study, the results of the other three experiments are all from the official technical reports of Qwen2.5-7B. Their functions are generally to measure and compare the performance of LLM models from various aspects. In this section, they represent the reasoning ability of the model itself that is opposite to the legal text quality capability proposed in this study. Therefore, the comparison of multiple benchmark results can more comprehensively show the performance changes of the model when affected by quantization.

\textbf{Score (Mean):} Our proposed metric for evaluating legal text quality.

\textbf{MMLU:} The Massive Multitask Language Understanding(MMLU) is a widely used benchmark for multi-domain knowledge understanding~\cite{hendrycks2020measuring}.

\textbf{C-Eval:} A Chinese evaluation suite covering multiple disciplines, widely adopted in LLM benchmarking~\cite{huang2023c}.

\textbf{IEVAL:} A task designed to assess instruction-following capabilities~\cite{svikhnushina2022ieval}.

\begin{table}[h]
 \centering
 \caption{Performance Comparison of Qwen2-7B-Instruct with Different Quantization Methods.}
 \label{tab:Quantization_1}

 \begin{tabular}{lcccc}
  \toprule
  \textbf{Model} & \textbf{Score (Mean)} & \textbf{MMLU} & \textbf{C-Eval} & \textbf{IEVAL} \\
  \midrule
  BF16 & 79.6 & 70.5 & 77.2 & 53.1 \\
  Int8 & 78.76 & 69.1 & 76.7 & 52.9 \\
  Int4 & 77.56 & 67.8 & 75.2 & 49.4 \\
  \bottomrule
  \end{tabular}
  \begin{tablenotes}  
  \footnotesize       
  \item All models are different Qwen2.5-7B-Instruct quantization version;
  \end{tablenotes}

\end{table}

The results indicate that moving from BF16 to Int8 only introduces a negligible performance drop across all benchmarks, confirming the effectiveness of Int8 quantization for efficiency-oriented deployment. However, the transition to Int4 leads to more noticeable degradation, particularly on knowledge-intensive tasks such as MMLU and C-Eval. This highlights the trade-off between efficiency and representational fidelity when adopting extreme quantization.

To statistically assess the impact of quantization, we first tested the normality of score distributions using the Shapiro–Wilk test, which examines whether a sample is drawn from a normally distributed population~\cite{shapiro1965analysis}. The results showed that the BF16 (Qwen2.5-7B) setting deviated significantly from normality ($p=0.0095$), whereas the Int8 ($p=0.1182$) and Int4 ($p=0.3089$) settings did not. Given this violation of the normality assumption for BF16, we employed the Wilcoxon signed-rank test, a non-parametric alternative to the paired $t$-test, to evaluate whether the median of paired differences between quantization settings differs from zero~\cite{wilcoxon1945individual}. 

Because three pairwise tests were performed, we applied multiple-comparison corrections to control the risk of inflated Type I error. Specifically: 
\begin{itemize}
    \item \textbf{Bonferroni correction}: Divides the overall significance level $\alpha=0.05$ by the number of tests ($m=3$), yielding an adjusted $\alpha \approx 0.0167$ \cite{bonferroni1936teoria,dunn1961multiple}.
    
    \item \textbf{Holm–Bonferroni correction}: Orders the $p$-values and sequentially tests them against increasingly lenient thresholds, offering more power while still controlling the family-wise error rate (FWER) \cite{holm1979simple}.
    
    \item \textbf{Benjamini–Hochberg procedure (BH)}: Controls the false discovery rate (FDR), making it more appropriate for exploratory comparisons \cite{benjamini1995controlling}.

    \item \textbf{Hodges–Lehmann (HL) estimator}: Provides the median of all pairwise differences as a robust estimate of the shift between conditions, with 95\% bootstrap confidence intervals (CI)~\cite{hodges1963estimators}.
\end{itemize}

The results are summarized in Table~\ref{tab:Quantization}. None of the pairwise comparisons reached statistical significance after any correction. However, HL estimates indicate small performance drops when reducing to Int4, although all CIs include zero. This suggests that Int8 quantization is essentially equivalent to BF16, while Int4 presents a possible risk of quality degradation that is not statistically confirmed.

\begin{table}[h]
 \centering
 \caption{Pairwise comparisons of quantization methods using the Wilcoxon signed-rank test. Reported are raw $p$-values, adjusted $p$-values under Bonferroni, Holm, and Benjamini–Hochberg corrections, as well as Hodges–Lehmann (HL) median differences with 95\% bootstrap confidence intervals.}
 \label{tab:Quantization}
 \begin{tabular}{lcccccc}
  \toprule
  \textbf{Comparison} & \textbf{Statistic} & \textbf{raw $p$-value} & \textbf{p$_{Bonf}$} & \textbf{p$_{Holm}$} & \textbf{q$_{BH}$} & \textbf{HL $\tilde{\Delta}$ [95\% CI]} \\
  \midrule
  BF16 vs Int8  & 844.0 & 0.601 & 1.000 & 0.601 & 0.601 & 0.87 [-1.83, 3.83] \\
  BF16 vs Int4  & 681.0 & 0.085 & 0.255 & 0.255 & 0.255 & 1.99 [-0.76, 4.95] \\
  Int8 vs Int4  & 741.0 & 0.200 & 0.601 & 0.400 & 0.300 & 1.39 [-1.48, 4.57] \\
  \bottomrule
 \end{tabular}
 \begin{tablenotes}  
  \footnotesize       
  \item Note: None of the comparisons reached significance under corrected $\alpha=0.0167$. HL estimates suggest small but non-significant performance decreases for Int4. 
 \end{tablenotes}
\end{table}

In summary, Int8 quantization offers a practical compromise between efficiency and accuracy for legal text quality evaluation, while Int4 should be applied with caution due to observed degradation trends that did not reach statistical significance.

\subsubsection{Effect of Context Length}

Following the release of standard model versions, many providers subsequently develop variants with extended context windows through various architectural modifications. These different context length versions typically exhibit varying API call costs and latency characteristics. This section investigates the potential impact of context length on text generation quality, using the Kimi model \cite{kimiteam2025kimik15scalingreinforcement} as an illustrative example with three context window variants (8k, 32k, and 128k). Kimi is particularly suitable for this role because it provides standardized multi-context versions within the same product line and timeframe, which allows us to examine context length effects while minimizing confounding changes in architecture or training data. The experimental results are presented in Table~\ref{tab:Max_number_1}. In this subsection, we specifically examine how context length impacts the quality of legal text generation.

\begin{table}[htbp]
 \centering
 \caption{Performance of Kimi models with different context lengths (8k, 32k, 128k).}
 \label{tab:Max_number_1}

 \begin{tabular}{lcccc}
  \toprule
  \textbf{Model} & \textbf{Mean} & \textbf{Std} & \textbf{Median} & \textbf{AdjScore} \\
  \midrule
  Kimi\_8k & 76.53 & 8.55 & 75.81 & 68.84 \\
  Kimi\_32k & 77.25 & 7.92 & 75.89 & 70.06 \\
  Kimi\_128k & 75.67 & 6.94 & 74.95 & 69.32 \\
  \bottomrule
 \end{tabular}

\end{table}

For the Kimi model series, we applied the same statistical procedures as introduced in the section Effect of Quantization, including the Wilcoxon signed-rank test with Bonferroni, Holm–Bonferroni, and Benjamini–Hochberg corrections, as well as Hodges–Lehmann (HL) effect size estimation. The results are summarized in Table~\ref{tab:Kimi}. None of the pairwise comparisons reached statistical significance under the corrected threshold ($\alpha=0.0167$). Specifically, the raw $p$-values were 0.503 (Kimi\_8k vs. Kimi\_32k), 0.365 (Kimi\_8k vs. Kimi\_128k), and 0.230 (Kimi\_32k vs. Kimi\_128k), all of which remained non-significant after multiple-comparison adjustments. HL estimates suggested a small positive shift favoring the 128k variant (median differences of about 1.0 point), but all 95\% confidence intervals included zero, indicating that these trends are not statistically conclusive. 

\begin{table}[h]
 \centering
 \caption{Pairwise comparisons of Kimi models with different context lengths using Wilcoxon signed-rank tests. Reported are raw $p$-values, adjusted $p$-values (Bonferroni and Holm), Benjamini–Hochberg $q$-values, and Hodges–Lehmann (HL) effect sizes with 95\% confidence intervals.}
 \label{tab:Kimi}
 \begin{tabular}{lcccccc}
  \toprule
  \textbf{Comparison} & \textbf{Statistic} & \textbf{raw $p$-value} & \textbf{p$_{Bonf}$} & \textbf{p$_{Holm}$} & \textbf{q$_{BH}$} & \textbf{HL $\tilde{\Delta}$ [95\% CI]} \\
  \midrule
  Kimi\_8k vs Kimi\_32k   & 824.0 & 0.503 & 1.000 & 0.730 & 0.503 & -0.37 [-3.49, 2.44] \\
  Kimi\_8k vs Kimi\_128k  & 792.0 & 0.365 & 1.000 & 0.730 & 0.503 & 0.99 [-1.93, 4.06] \\
  Kimi\_32k vs Kimi\_128k & 752.0 & 0.230 & 0.691 & 0.691 & 0.503 & 1.27 [-1.34, 4.05] \\
  \bottomrule
 \end{tabular}
 \begin{tablenotes}  
  \footnotesize       
  \item None of the comparisons reached significance under corrected $\alpha=0.0167$. HL estimates suggest small but non-significant advantages for the 128k variant.
 \end{tablenotes}
\end{table}

In summary, context length variation from 8k to 128k tokens did not yield statistically significant quality differences. The 128k variant exhibited slightly higher HL estimates, but with wide confidence intervals overlapping zero. These findings suggest that context length in this range has negligible statistical impact on text quality, and practical deployment should instead be guided by economic considerations (e.g., API costs) and task requirements (e.g., necessary input size).

\subsubsection{Effect of Model Type (Reasoning vs Base)}

\begin{figure}[ht]
  \centering
  \includegraphics[width=0.9\textwidth]{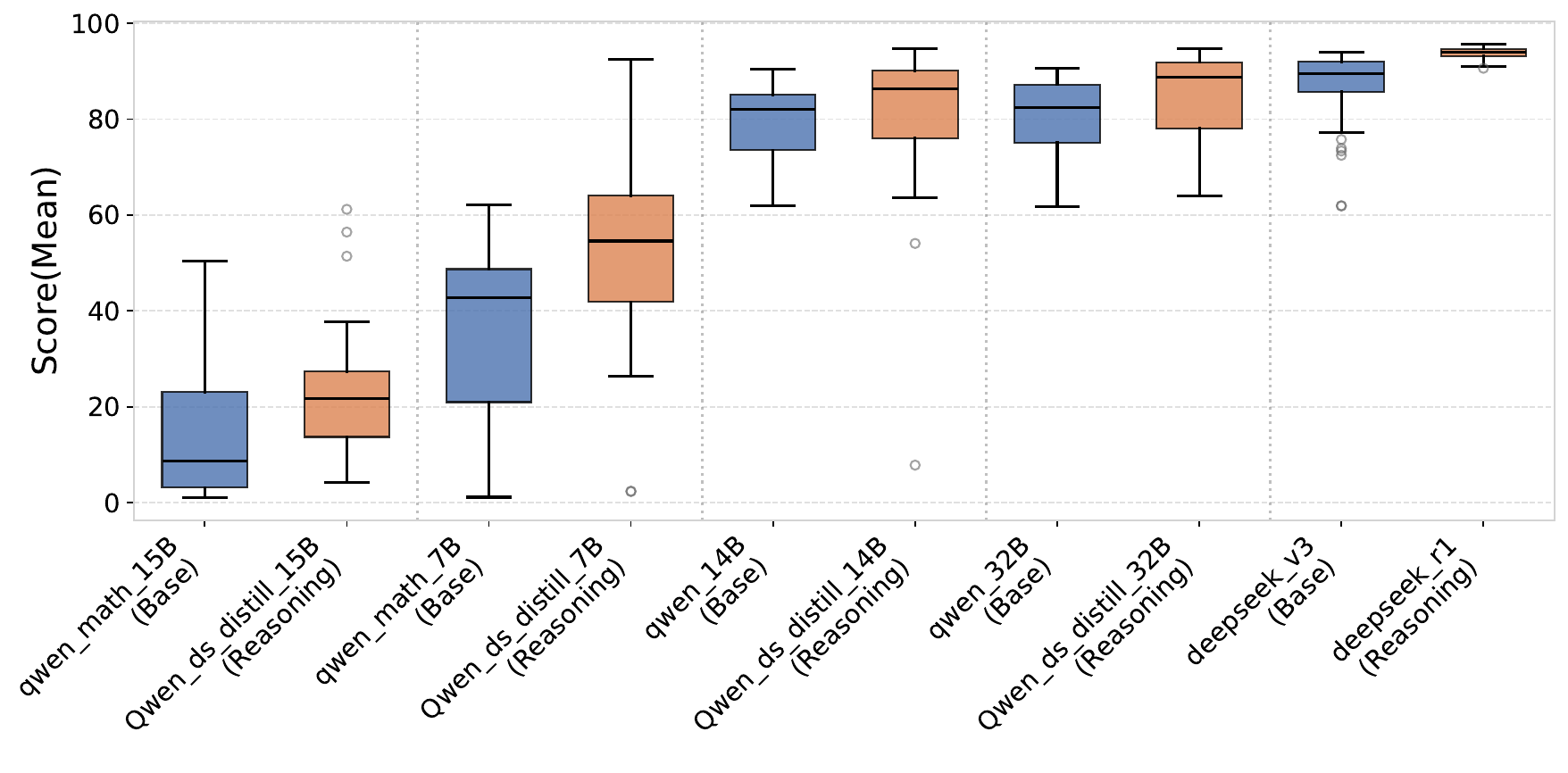}
  \caption{Performance Comparison Between Reasoning Models and Base Models: Evaluating Score (Mean) improvement of reasoning models compared to their base counterparts.}
  \label{fig:fig_dis_f4_Reasoning_Base}
\end{figure}

Reasoning-enhanced models have seen early adoption in real-world applications beyond academia, particularly in domains such as legal practice \cite{schwarcz2025ai} and healthcare \cite{temsah2025deepseek}. Meanwhile, general reasoning models also demonstrate stronger overall performance \cite{deepseekai2025deepseekr1incentivizingreasoningcapability}. The purpose of this section is to evaluate whether the advantages of reasoning models in terms of reasoning capability also extend to \emph{text quality}. Four control groups were selected: two reasoning models distilled from Qwen's mathematical model, one reasoning model distilled from Qwen's instruction model, and the R1 model released by Deepseek. In this subsection, we specifically examine whether reasoning-oriented architectures yield measurable improvements in \emph{text quality} compared with their base counterparts. The results are shown in Figure~\ref{fig:fig_dis_f4_Reasoning_Base}.

To verify the statistical significance of these observations, we conducted paired non-parametric tests (Wilcoxon signed-rank), consistent with the methodology described in the previous sections. For each base–reasoning pair, we also report the Hodges--Lehmann median paired difference with a 95\% bootstrap confidence interval. Unlike the quantization and context-length analyses where three-way comparisons required Bonferroni correction ($\alpha=0.0167$), here we only consider independent two-group comparisons. Therefore, we adopt a significance threshold of $\alpha=0.05$, while also reporting Holm-adjusted $p$-values for completeness. The full results are summarized in Table~\ref{tab:reasoning_base_stats}.

\begin{table}[ht]
\centering
\caption{Pairwise comparisons of reasoning models versus their base counterparts using Wilcoxon signed-rank tests. Reported are raw $p$-values, multiple-comparison adjusted $p$-values, and Hodges--Lehmann (HL) effect sizes with 95\% confidence intervals.}
\label{tab:reasoning_base_stats}
\begin{tabular}{lcccccc}
\toprule
\textbf{Comparison} & \textbf{Statistic} & \textbf{raw $p$-value} & \textbf{p$_{Bonf}$} & \textbf{p$_{Holm}$} & \textbf{q$_{BH}$} & \textbf{HL $\tilde{\Delta}$ [95\% CI]} \\
\midrule
Qwen2.5-1.5B & 379.0 & 0.0001 & 0.0005 & 0.0002 & 0.0002 & +7.98 [3.80, 14.52] \\
Qwen2.5-7B   & 289.0 & $<0.0001$ & $<0.0001$ & $<0.0001$ & $<0.0001$ & +18.76 [6.52, 24.45] \\
Qwen2.5-14B  & 619.0 & 0.0293 & 0.147 & 0.0293 & 0.0147 & +3.62 [1.30, 6.92] \\
Qwen2.5-32B  & 466.0 & 0.0009 & 0.0045 & 0.0019 & 0.0019 & +5.61 [2.88, 7.19] \\
Deepseek     & 16.0  & $<0.0001$ & $<0.0001$ & $<0.0001$ & $<0.0001$ & +3.89 [3.11, 5.69] \\
\bottomrule
\end{tabular}
\begin{tablenotes}
\footnotesize
\item All comparisons were significant at $\alpha=0.05$. Unlike the quantization and context-length analyses where Bonferroni correction ($\alpha=0.0167$) was applied, here only independent two-group comparisons were considered. We therefore adopt $\alpha=0.05$ as the reference threshold, while reporting Bonferroni-, Holm-, and Benjamini--Hochberg–adjusted values for completeness. 
\item Abbreviations: ``Qwen2.5-1.5B/7B/14B/32B'' denote \emph{Qwen2.5-Math-$X$ vs. Qwen2.5-Distill-$X$} pairs; ``Deepseek'' denotes \emph{Deepseek-V3 vs. Deepseek-R1}.
\end{tablenotes}
\end{table}

The results confirm a consistent advantage of reasoning-enhanced models across all scales. For smaller models (7B and 15B), the performance improvements are both statistically significant and large in magnitude, with HL effect sizes exceeding +8 points. At larger scales (14B and 32B), the improvements remain statistically significant, though the effect sizes diminish, consistent with our scaling-law analysis indicating that text quality saturates around the 14B parameter level. The Deepseek comparison shows an exceptionally small Wilcoxon statistic ($W=16$), indicating that nearly all paired samples favor Deepseek-V3 over Deepseek-R1, thus confirming a very robust advantage. Overall, these findings highlight that reasoning-specific distillation confers consistent text-quality benefits, but with diminishing marginal returns as model capacity increases.

\subsection{Engineering Trade-offs}
% 实际部署中的性能–成本平衡 %

When deploying large language models in production environments, performance is only one side of the equation; operational costs and efficiency constraints are equally critical. In contrast to Section~6.1, which examined intrinsic factors influencing model quality, this section focuses on external trade-offs that emerge in deployment. We discuss three representative dimensions: (i) model scale, (ii) output text length, and (iii) API pricing. By analyzing cost–quality frontiers and resource implications, we identify the configurations that offer the most favorable balance between performance and efficiency in real-world legal applications.

\subsubsection{Trade-off in Model Scale}

\begin{figure*}[ht]
  \centering
  \includegraphics[width=0.9\textwidth]{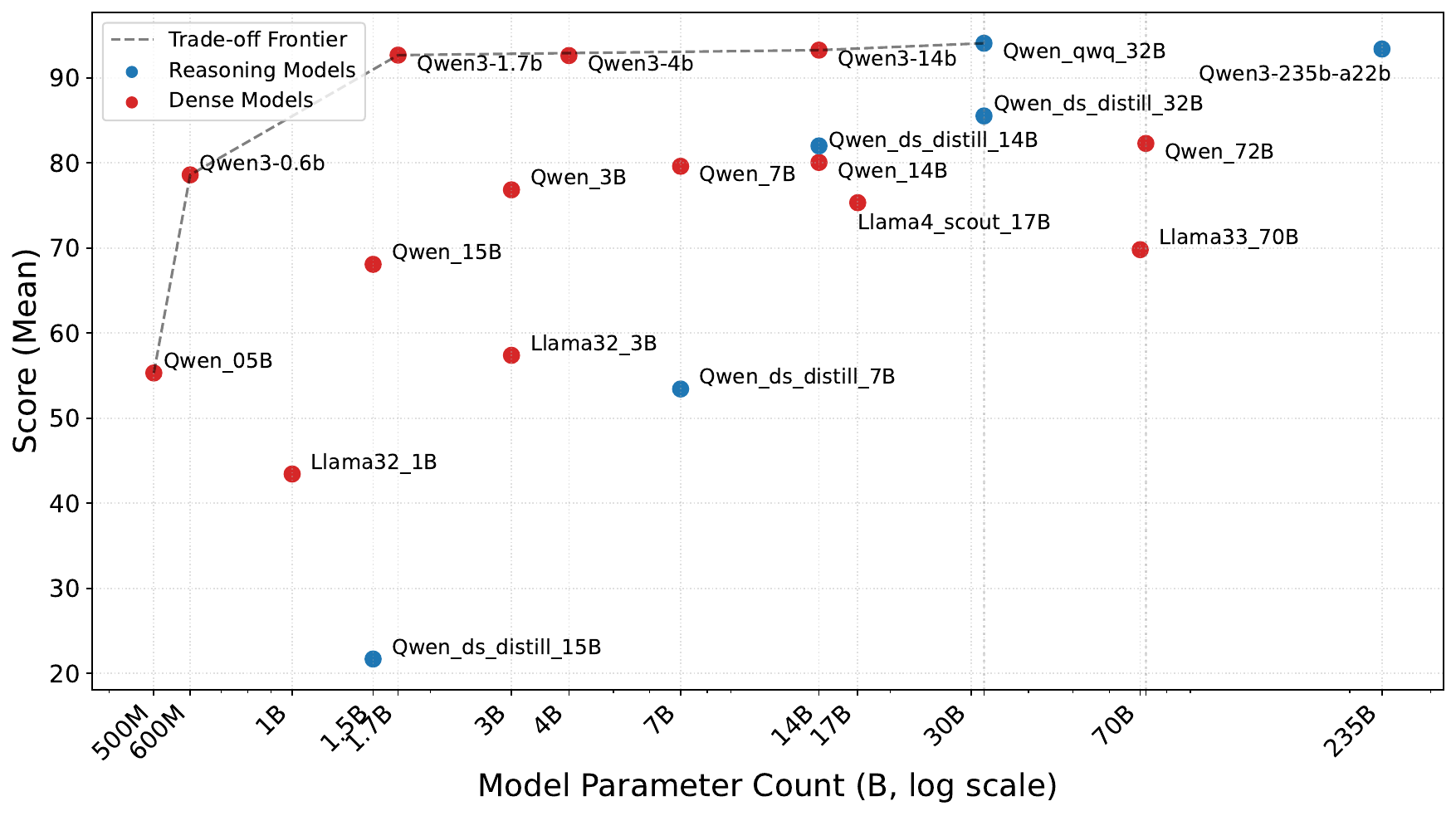}
  \caption{Scatter plot of Open-Source LLMs: Comparative Analysis of Qwen3, Qwen2.5, and Llama.}
  \label{fig:fig_dis_f3_Score_paraments}
\end{figure*}

Beyond the scaling-law analysis in Section~6.1, which highlighted diminishing returns of very large models, we here examine the trade-offs between parameter size and cost–performance efficiency. In total, 42 open-source models were analyzed; for clarity, Figure~\ref{fig:fig_dis_f3_Score_paraments} presents 20 representative models drawn from four major families (Qwen3, Qwen2.5, Deepseek-distill, and Llama3). Models lying on the trade-off frontier are highlighted as the most cost-efficient choices.

Three insights emerge from this analysis.  
(i) Qwen3-series models consistently dominate the frontier, offering 12–18\% higher legal text quality than peers of comparable parameter size.  
(ii) Comparing the Qwen2.5 series with its reasoning-oriented counterparts (Qwen2.5-reasoning and Deepseek-distill) reveals scale-dependent differences: the 1.5B variant (abbreviated as “15B” in the figure) underperforms its dense baseline, suggesting that this scale is insufficient to sustain stable reasoning capability. In contrast, at 32B and above, reasoning-optimized models clearly outperform dense architectures in parameter efficiency.  
(iii) Within the Qwen2.5 family, the 7B–14B range emerges as the optimal cost–performance sweet spot, aligning with the scaling-law observations, whereas in the Qwen3 family the optimum shifts toward much smaller models, with the 1.7B variant showing the best balance.

Taken together, these findings suggest three practical guidelines for deployment: (1) balance capability ceilings against throughput costs, (2) prioritize architecturally efficient designs over raw parameter scaling, and (3) concentrate deployment within the identified optimal parameter ranges.

\subsubsection{Trade-off in Output Text Length}
\label{sec:Trade-off_in_Output_Text_Length}

\begin{figure}[ht]
  \centering
  \includegraphics[width=0.9\textwidth]{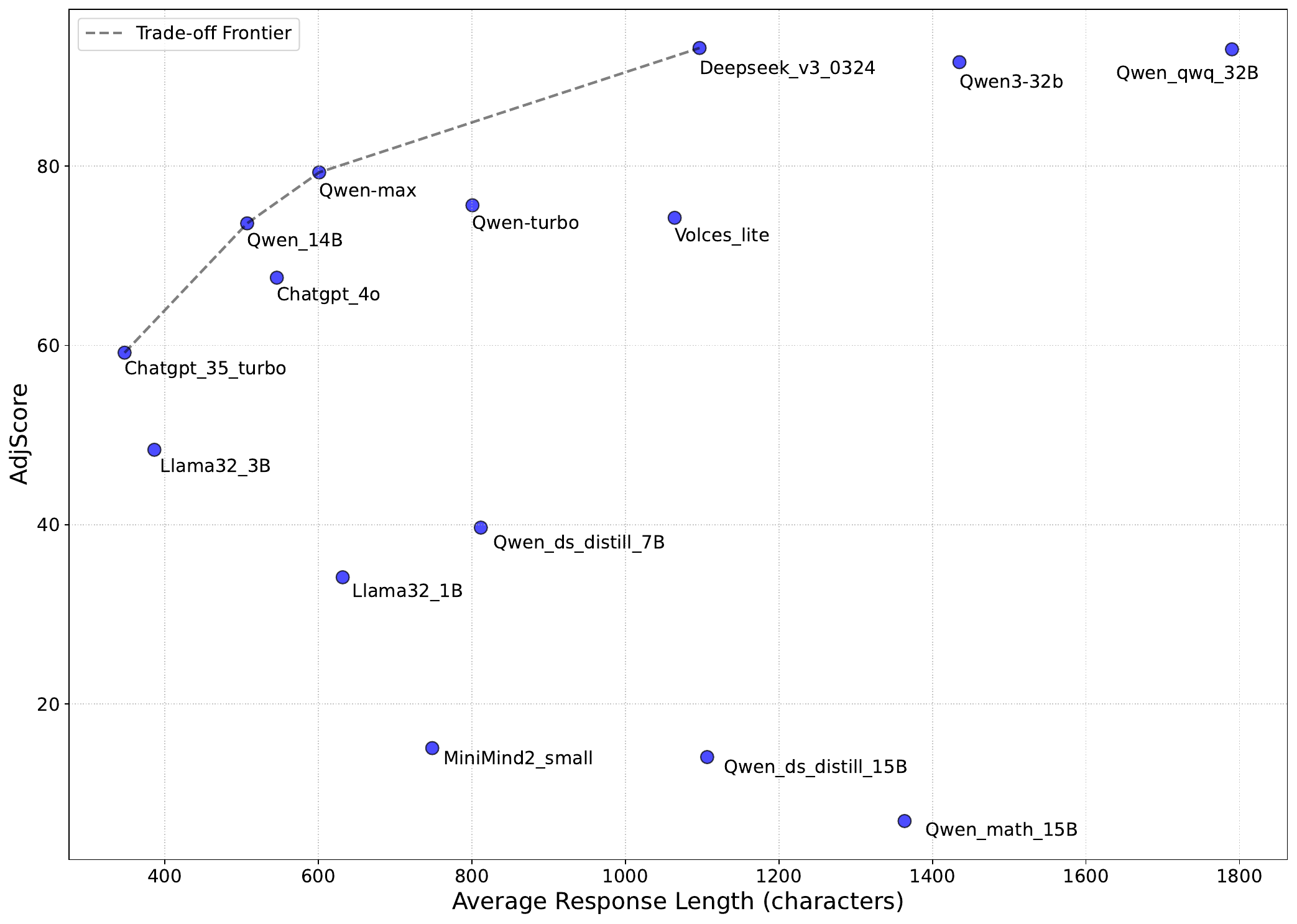}
  \caption{Relationship between output length and quality: scatter plot showing independent variation between output length and AdjScore.}
  \label{fig:fig_dis_f5_output_AdjCV}
\end{figure}

We next examine whether output length, which directly increases inference cost, provides consistent benefits in text quality (measured by AdjScore). Figure~\ref{fig:fig_dis_f5_output_AdjCV} plots the average output length against mean AdjScore for each model.

To complement the visualization, we conducted correlation analyses using Spearman’s rank correlation~\cite{spearman1904TheProof} and Kendall’s $\tau_b$~\cite{kendall1938measure}. Confidence intervals for Spearman’s $\rho$ were estimated via Fisher’s $z$ transformation and bootstrap resampling.

The results (Table~\ref{tab:output_length_corr}) show no statistically significant association between output length and AdjScore (Spearman’s $\rho=0.18$, $p=0.52$; Kendall’s $\tau_b=0.10$, $p=0.63$). Robustness checks, including outlier removal, restriction to the Qwen family, and stratified analyses by response length, consistently yielded non-significant outcomes. Quartile-based analyses produced higher coefficients (up to $\rho=0.8$), but these were based on only $n=4$ observations and thus lack reliability.

\begin{table}[htbp]
\centering
\caption{Correlation between output length and AdjScore across models (primary and robustness checks). Reported are correlation coefficients and raw $p$-values.}
\label{tab:output_length_corr}
\begin{tabular}{lcccc}
\toprule
\textbf{Analysis} & \textbf{Test} & \textbf{n} & \textbf{Correlation} & \textbf{raw $p$-value} \\
\midrule
Primary (all models) & Spearman & 15 & 0.179 & 0.524 \\
                     & Kendall  & 15 & 0.105 & 0.627 \\
Outlier removal (score) & Spearman & 15 & 0.179 & 0.524 \\
Outlier removal (length) & Spearman & 15 & 0.179 & 0.524 \\
Qwen family only        & Spearman &  8 & 0.238 & 0.570 \\
Median split: short     & Spearman &  8 & 0.024 & 0.955 \\
Median split: long      & Spearman &  7 & 0.179 & 0.702 \\
Quartile split: Q1      & Spearman &  4 & 0.600 & 0.400 \\
Quartile split: Q4      & Spearman &  4 & 0.800 & 0.200 \\
\bottomrule
\end{tabular}
\footnotesize
\item For the primary Spearman analysis, the 95\% bootstrap confidence interval was [$-0.48$, $0.73$]. 
\item Adjusted $p$-values (Holm and Benjamini--Hochberg) were omitted for brevity, as all were non-significant (Holm = 1.0, BH = 0.835).
\end{table}

Since inference cost is directly proportional to the number of output tokens under standard billing schemes ($( \text{input tokens} + \text{output tokens}) \times \text{unit price}$), longer outputs also incur higher costs in addition to the quality trade-offs discussed above. This simplified formulation assumes a single unit price, whereas in practice many providers apply different rates for input and output tokens, typically with higher costs for output.

Overall, these findings demonstrate that output length is not a reliable predictor of legal text quality. High-quality answers can be concise (upper-left region), while some very long outputs correspond to lower quality (lower-right region). This suggests that coherence and reasoning ability matter more than verbosity, and increasing output length primarily inflates inference cost without guaranteeing improvements in legal text quality.

\subsubsection{Trade-off in API Price}

Building on Section~\ref{sec:Trade-off_in_Output_Text_Length}, which established that output length inflates inference cost, we now focus on the second major cost factor: the unit price of tokens. Among providers such as OpenAI, Anthropic, Qwen, Deepseek, and Kimi, pricing schemes are consistently based on token counts, and prior studies confirm that token-based billing directly shapes the sustainability and profitability of large-scale LLM adoption~\cite{shekhar2024optimizingcostsllmusage,nandagopaltokens}. All API prices reported in this section were retrieved from official provider pricing pages on May 20, 2025. To maintain consistency and simplify comparisons, we use the unit price of output tokens as the basis of cost evaluation, disregarding potential asymmetries between input and output token pricing.

\begin{figure}[ht]
  \centering
  \includegraphics[width=0.9\textwidth]{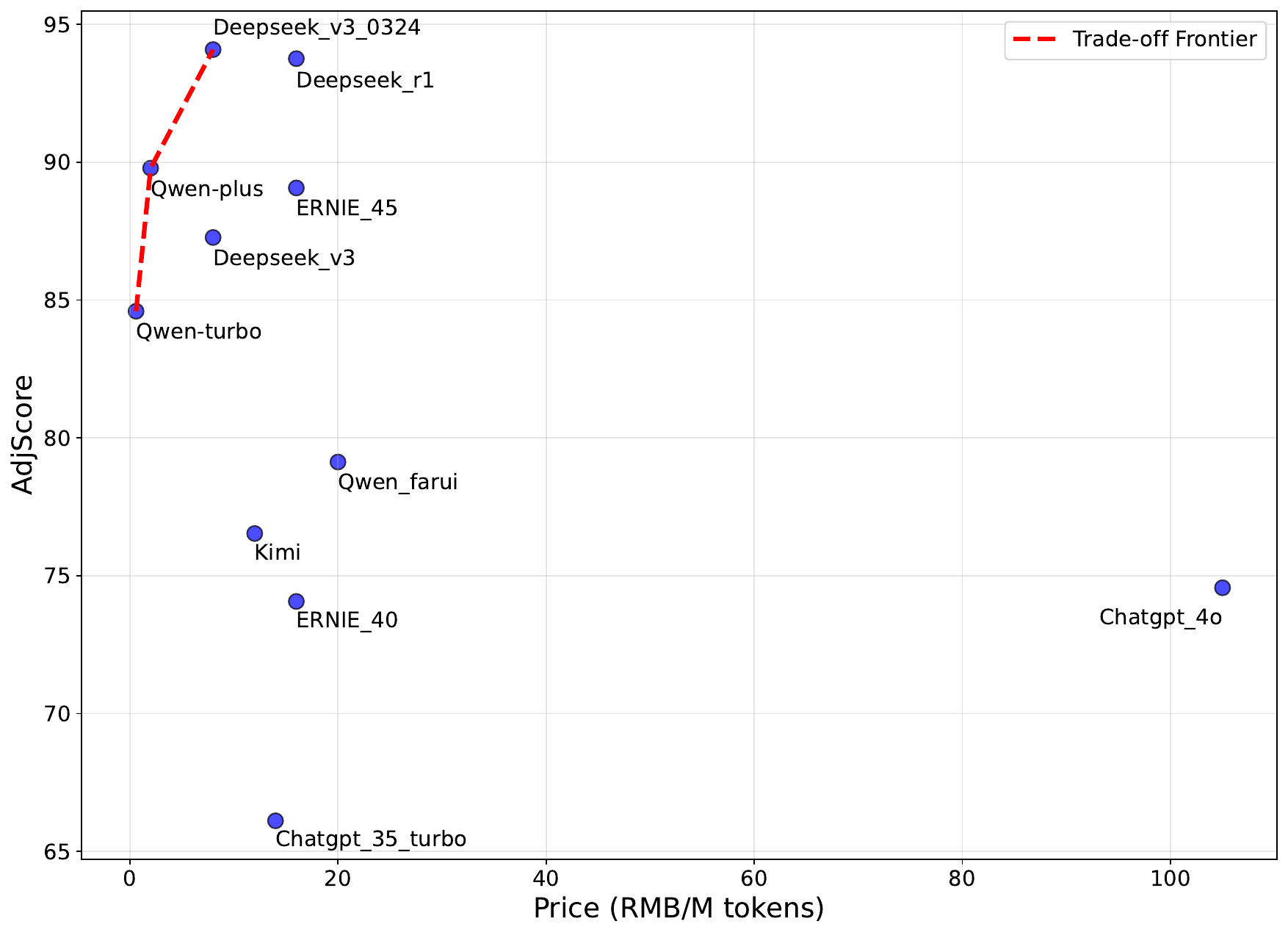}
  \caption{Relationship between token unit price and AdjScore across major LLM APIs.}
  \label{fig:fig_dis_f6_Price_AdjCV}
\end{figure}

Figure~\ref{fig:fig_dis_f6_Price_AdjCV} plots unit token price against AdjScore. ChatGPT-4o appears as a clear outlier due to its premium tier, whereas the trade-off frontier is shaped by three cost-efficient options: Qwen-Turbo, Volces-Pro, and Deepseek-v3-0324. Importantly, API pricing is subject to temporal variability driven by promotions, distribution channels, and model updates, necessitating periodic reevaluation.

From a deployment perspective, three guidelines emerge:  
(i) API prices should be reassessed regularly given their volatility;  
(ii) cost–quality frontiers help identify efficient choices, such as Qwen-Turbo for maximum cost efficiency or Deepseek-v3-0324 when higher-quality guarantees are critical;  
(iii) enterprises should base model selection on holistic cost–performance trade-offs rather than isolated considerations of price or accuracy.

\section{Threats to Validity}

To systematically assess the robustness and potential limitations of our findings, we organize this section using the standard taxonomy of threats to validity widely adopted in software engineering and machine learning research~\cite{cook1979quasi, wohlin2012experimentation}, covering internal, external, construct, and conclusion validity.

\subsection{Internal Validity}
Our framework relies on annotated training data generated through a structured pipeline of student annotators and expert oversight. Although multiple quality-control measures were applied, residual annotator bias or inconsistency may still influence score distributions. In addition, annotation granularity for top-tier outputs remains limited, which could reduce discrimination among high-quality models. Although this issue also overlaps with construct validity, we emphasize it here as it arises directly from the practical annotation process.

\subsection{External Validity}
The present instantiation of the framework is tailored to the legal domain. While this ensures domain relevance, it may limit generalizability to other professional contexts such as medicine or technical documentation. Beyond domain generalization, we also examined cross-linguistic applicability through a translation-based zero-shot validation on English data. Specifically, queries and model-generated answers from the Chinese dataset were translated into English using commercial APIs, and the English pairs were evaluated under identical conditions. Statistical comparisons between Chinese and English evaluations (paired $t$-test and Wilcoxon signed-rank) revealed no significant differences, with effect sizes close to zero. These findings suggest consistent assessments across Chinese and English inputs, providing preliminary evidence for cross-linguistic applicability. Detailed results, including statistical tests and diagnostic visualizations, are reported in Appendix~C.

\subsection{Construct Validity}
Our evaluation relies on a composite AdjScore as the primary proxy for text quality. While this metric integrates multiple dimensions, it may not fully capture all aspects of legal reasoning or professional writing. Moreover, the use of a Sigmoid activation function constrains scores within $(0,1)$ (rescaled to $(0,100)$). In practice, scores saturate around 96, reflecting data distribution and training effects rather than a theoretical ceiling. This saturation may restrict differentiation among top-performing models such as Qwen3 and Deepseek, posing a construct validity threat.

\subsection{Conclusion Validity}
Our comparative analyses employed non-parametric statistical tests (e.g., Wilcoxon signed-rank) with bootstrap confidence intervals and multiple-testing corrections to mitigate spurious findings. Nevertheless, statistical conclusions remain contingent on dataset size, model coverage, and potential annotator noise. While the tests indicate significant improvements in several comparisons, caution is warranted in interpreting marginal effects, particularly when results are close to conventional significance thresholds and may not correspond to practically meaningful gains.

Overall, these threats do not undermine the utility of the framework but highlight directions for future work to strengthen robustness, scalability, generalizability, and interpretability.

\section{Future Directions}

Building on the identified threats to validity, several directions emerge to address current limitations and extend the framework’s applicability. 

First, to mitigate the current legal-domain specialization, future work should extend the framework to multiple professional contexts, such as medicine, finance, and technical documentation. Constructing multi-domain corpora and training evaluation models on diversified data may enhance external validity and ensure broader applicability. However, scores cannot be assumed to be directly comparable across domains, as each professional context has distinct quality criteria. Domain-specific calibration or retraining will therefore remain essential to maintain validity. 

Second, the current use of a Sigmoid activation restricts the dynamic range of scores, leading to saturation effects around 96. Alternative formulations, such as linear or piecewise-linear activations or dynamic rescaling strategies, may expand the effective range. This would improve differentiation among top-performing models and reduce ceiling effects. 

Third, enhancing the granularity and consistency of annotations represents another key direction. Hierarchical rating scales can capture more fine-grained distinctions among outputs, while multi-stage expert validation can reinforce consistency and reduce annotator bias. Such refinements would strengthen internal validity while preserving annotation efficiency. 

Finally, integrating adaptive quality metrics, domain-specific expert review, and semi-automated annotation assistance could improve both reliability and scalability. These developments would enable application to industrial-scale evaluation tasks and allow more nuanced quality discrimination in practice. 

Overall, these directions highlight the potential to evolve the current framework into a more robust, generalizable, and fine-grained evaluation system that serves the needs of both diverse professional domains and large-scale industrial deployment, while accommodating continuous advances in LLM architectures and evaluation criteria.

\section{Conclusion}

This study introduces a quantitative framework for evaluating the quality of legal texts generated by large language models through regression analysis, thereby addressing the lack of standardized assessment metrics for domain-specific applications. A systematic evaluation of 49 models across diverse architectures yields three principal findings. First, legal text quality exhibits scaling characteristics that diverge from those observed in mathematical reasoning, showing clear diminishing returns beyond 7B parameters, with negligible gains past 14B. Second, engineering optimizations, including quantization strategies and extensions of context windows, show no statistically significant effect on legal text quality ($p > 0.0167$), supporting the feasibility of cost-efficient deployment configurations. Third, reasoning-oriented models consistently outperform their base counterparts despite architectural similarity, underscoring the benefits of reasoning-specific training and offering new insights for domain-specific evaluation of large language models.

The developed multidimensional assessment model provides practitioners with actionable guidance for model selection through rigorous validation in the legal domain and trade-off analysis, highlighting the Qwen3 series as the current leaders in parameter efficiency. These findings challenge conventional scaling paradigms by demonstrating that inherent limitations in training data quality cannot be overcome by parameter scaling alone, while also offering a new perspective for domain-specific evaluation methodologies. 

Future research should prioritize extending the framework to multidisciplinary datasets, developing dynamic activation functions to mitigate saturation effects, refining annotation granularity and consistency, and establishing industry-standard benchmarks. Together, these directions can advance both methodological foundations and practical deployment of evaluation systems in diverse professional contexts.\\

\noindent\textbf{Funding} No funding was received for this study. \\ 

\noindent\textbf{Informed consent} Not applicable. \\  

\noindent\textbf{Data availabilty} The data that support the findings of this study are not openly available due to reasons of sensitivity and are available from the corresponding author upon reasonable request. \\
    
\noindent\textbf{Author contributions}  Y.L. conceived the main idea, designed and implemented the system, conducted the experiments, and drafted the manuscript. G.W. contributed to refining the main idea, offered advice and suggestions, helped improve the manuscript, and supervised the project. All authors reviewed the manuscript.

\section*{Declarations}

\noindent\textbf{Conflict of interest} The authors declare no competing interests. \\

%\bibliography{sn-bibliography}
%% BioMed_Central_Bib_Style_v1.01

\appendix

\clearpage 

\section*{Appendix A  Comprehensive Ranking of Experimental Results} 

This appendix provides the full ranking of all 49 evaluated models to complement the representative subsets reported in the main text. The table below lists each model together with its mean score, standard deviation, and the stability-adjusted score (AdjScore). 

As described in Section~\ref{sec:experiments}, the mean score reflects the average quality of model outputs across the validation set, while the standard deviation captures the variability of performance. The stability-adjusted score integrates both aspects into a single measure, penalizing models that exhibit unstable performance despite high average quality. This metric enables more reliable comparisons among models with similar mean scores but different levels of consistency. 

The results reveal clear stratification across the evaluated models. The top tier is dominated by Deepseek-v3-0324, Qwen-QWQ-32B, and Deepseek-R1, all achieving AdjScores above 92, reflecting both high quality and stable performance. Mid-tier models such as ERNIE 4.5, Volces-Pro, and Qwen2.5-7B maintain strong averages but with moderately higher variance, resulting in AdjScores in the 70–85 range. The bottom tier, including smaller-scale or distillation-specialized variants such as Qwen2.5-0.5B and Qwen-Math-15B, demonstrates limited performance and higher instability, reflected in AdjScores below 50. 

For transparency, all models are listed in Table~\ref{tab:model_performance}, ensuring that readers can fully examine the complete experimental outcomes and verify the comparative claims made in the main paper.

\begin{longtable}{lrrr}
  \caption{Model Performance Comparison.} \label{tab:model_performance} \\
  \toprule
  \textbf{Model} & \textbf{Mean} & \textbf{Std} & \textbf{AdjScore} \\
  \midrule
  \endfirsthead

  \multicolumn{4}{c}%
  {{\bfseries Table \thetable\ (continued)}} \\
  \toprule
  \textbf{Model} & \textbf{Mean} & \textbf{Std} & \textbf{AdjScore} \\
  \midrule
  \endhead

  \bottomrule
  \endfoot

  Deepseek\_v3\_0324     & 94.09 & 0.93 & 93.16 \\
  Qwen\_qwq\_32B       & 94.06 & 1.06 & 93.01 \\
  Deepseek\_r1       & 93.76 & 1.23 & 92.54 \\
  Qwen3-14b        & 93.25 & 1.17 & 92.09 \\
  ERNIE\_X1         & 92.95 & 1.12 & 91.85 \\
  Qwen3-235b-a22b     & 93.38 & 1.65 & 91.76 \\
  Qwen3-32b        & 93.04 & 1.48 & 91.58 \\
  Qwen3-30b-a3b      & 92.87 & 1.62 & 91.28 \\
  Qwen3-1.7b        & 92.66 & 2.06 & 90.65 \\
  Qwen3-4b         & 92.61 & 2.60 & 90.08 \\
  Qwen-plus        & 89.78 & 5.77 & 84.36 \\
  ERNIE\_45         & 89.07 & 4.99 & 84.34 \\
  Volces\_pro        & 89.59 & 5.73 & 84.21 \\
  Deepseek\_v3       & 87.27 & 7.08 & 80.72 \\
  Qwen-max         & 84.65 & 5.72 & 79.29 \\
  Qwen\_ds\_distill\_32B   & 85.51 & 7.74 & 78.42 \\
  Qwen\_72B         & 82.29 & 6.73 & 76.07 \\
  Qwen-turbo        & 84.59 & 10.04 & 75.62 \\
  Volces\_lite       & 84.84 & 12.14 & 74.22 \\
  Qwen\_14B         & 80.05 & 7.01 & 73.60 \\
  Qwen\_32B         & 80.18 & 8.05 & 72.87 \\
  Qwen\_7B         & 79.60 & 7.60 & 72.66 \\
  Qwen\_farui        & 79.12 & 7.71 & 72.10 \\
  Qwen\_7B\_int8       & 78.76 & 7.84 & 71.64 \\
  Qwen\_ds\_distill\_14B   & 81.99 & 12.98 & 70.79 \\
  Qwen\_7B\_int4       & 77.56 & 7.55 & 70.68 \\
  Qwen\_3B         & 76.83 & 7.15 & 70.29 \\
  Kimi\_32k         & 77.24 & 7.92 & 70.06 \\
  Kimi\_128k        & 75.67 & 6.94 & 69.32 \\
  Kimi           & 76.53 & 8.55 & 68.84 \\
  Llama4\_maverick-17B   & 75.77 & 8.91 & 67.80 \\
  Chatgpt\_4o        & 74.56 & 7.75 & 67.54 \\
  ERNIE\_40         & 74.07 & 7.72 & 67.08 \\
  Qwen3-0.6b        & 78.59 & 14.49 & 66.36 \\
  Llama4\_scout\_17B     & 75.32 & 10.59 & 66.03 \\
  ERNIE\_speed       & 70.10 & 9.11 & 62.04 \\
  Llama33\_70B       & 69.80 & 9.62 & 61.35 \\
  Chatgpt\_35\_turbo     & 66.11 & 7.74 & 59.18 \\
  Qwen\_15B         & 68.08 & 11.15 & 58.49 \\
  Llama32\_3B        & 57.38 & 10.72 & 48.35 \\
  Qwen\_05B\_int8      & 55.68 & 11.64 & 46.05 \\
  Qwen\_05B         & 55.31 & 12.43 & 45.16 \\
  Qwen\_ds\_distill\_7B    & 53.43 & 18.53 & 39.67 \\
  Qwen\_05B\_int4      & 45.81 & 14.06 & 35.05 \\
  Llama32\_1B        & 43.44 & 11.84 & 34.13 \\
  Qwen\_math\_7B       & 35.84 & 17.41 & 24.13 \\
  MiniMind2\_small     & 23.23 & 12.55 & 15.08 \\
  Qwen\_ds\_distill\_15B   & 21.69 & 11.73 & 14.08 \\
  Qwen\_math\_15B      & 13.58 & 12.99 & 6.94 \\
\end{longtable}

\clearpage
\section*{Appendix B  Illustrative Examples of Complete Evaluation Cases}

This appendix presents three representative cases to illustrate how the LegalEval-Q framework evaluates model outputs across the five components: \textbf{query, answer, comment, conclusion, and score}. The examples cover different levels of text quality, corresponding to low, medium, and high scores. For clarity, the detailed contents of all five components are shown directly inside the figures.  

\subsection*{B.1 Negative Example (Score: 0)}
\begin{figure}[h]
  \centering
  \includegraphics[width=0.85\textwidth]{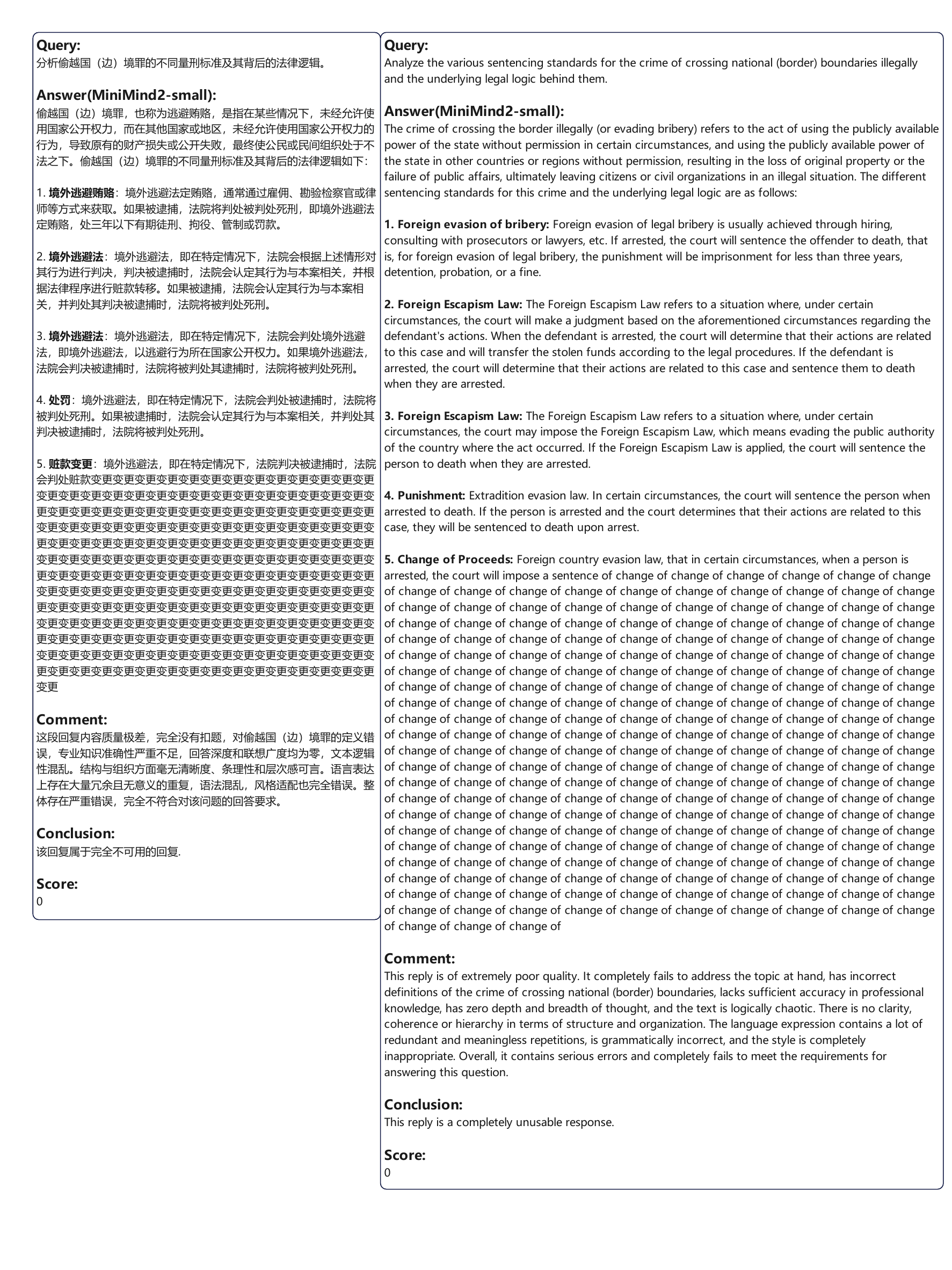}
  \caption{Negative example (score: 0) generated by the MiniMind2-small model, showing the complete five-part evaluation (query, answer, comment, conclusion, score).}
  \label{fig:fig_append_data_example_minimind2_0}
\end{figure}

This case demonstrates a very low-quality response generated by the MiniMind2-small model. The reply is completely irrelevant to the query, lacks professional accuracy, and contains severe logical and linguistic flaws. As shown in Figure~\ref{fig:fig_append_data_example_minimind2_0}, the evaluation framework assigns a score of~0, with comments and conclusions highlighting the critical deficiencies across all dimensions.  

\subsection*{B.2 Medium Example (Score: 58)}
\begin{figure}[h]
  \centering
  \includegraphics[width=0.85\textwidth]{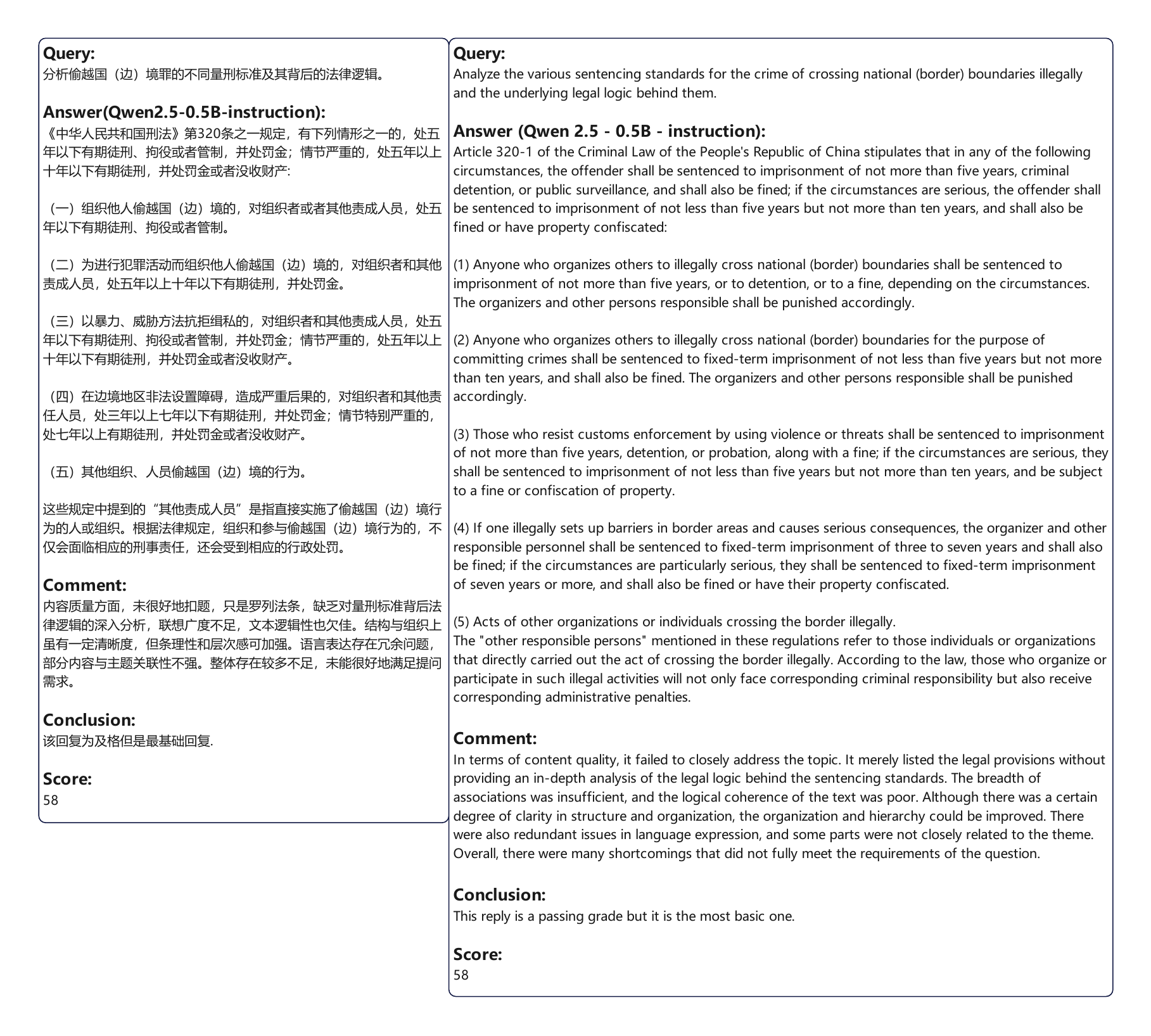}
  \caption{Medium-scoring example (score: 58), illustrating a case with partial relevance and moderate reasoning. The full five components are displayed (query, answer, comment, conclusion, score).}
  \label{fig:fig_append_data_example_qwen05B_58_zipped}
\end{figure}

This case presents a medium-quality response with partial relevance and moderate reasoning. As illustrated in Figure~\ref{fig:fig_append_data_example_qwen05B_58_zipped}, the evaluation framework assigns a score of~58. The results show that while the response contains some correct elements, it suffers from insufficient depth and limited coherence, which are reflected in the associated comments and conclusion.  

\subsection*{B.3 Positive Example (Score: 96)}
\begin{figure}[h]
  \centering
  \includegraphics[width=0.85\textwidth]{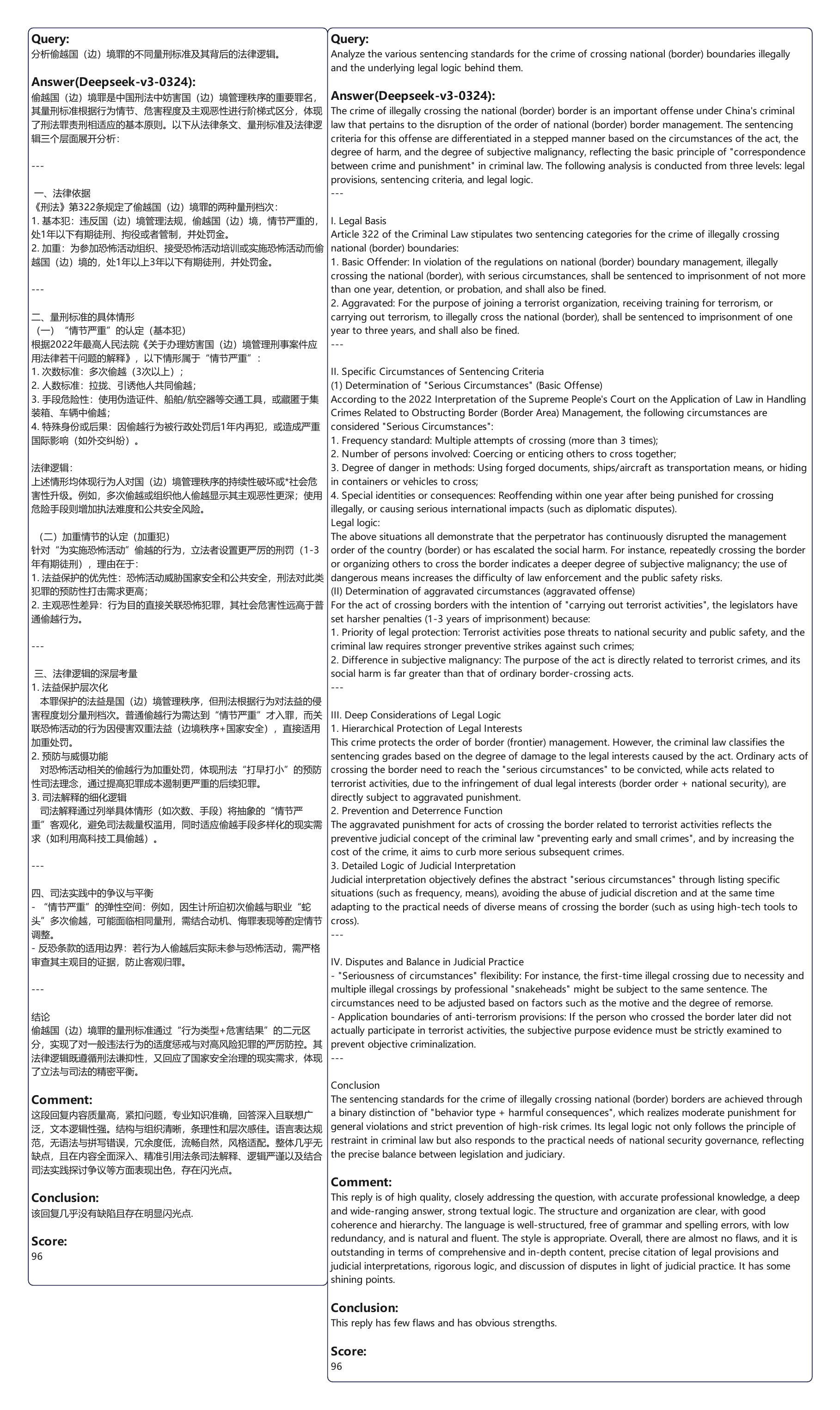}
  \caption{Positive example (score: 96) generated by the Deepseek-v3-0324 model, illustrating an almost flawless evaluation across all five components (query, answer, comment, conclusion, score).}
  \label{fig:fig_append_data_example_deepseekv3_0324_96_zipped}
\end{figure}

This case shows a high-quality response generated by the Deepseek-v3-0324 model. The answer is closely related to the query, employs accurate professional terminology, and demonstrates strong logical organization. As displayed in Figure~\ref{fig:fig_append_data_example_deepseekv3_0324_96_zipped}, the evaluation framework assigns a score of~96, with comments and conclusion confirming its comprehensive coverage, stylistic appropriateness, and rigorous reasoning.  

\clearpage
\section*{Appendix C Cross-Language Validation of LegalEval-Q}
\label{sec:Appendix_Cross-Language}

\subsection{Workflow of Cross-Language Evaluation}
To evaluate the cross-linguistic robustness of the LegalEval-Q framework, we implemented a translation-based validation procedure. Specifically, the queries in the Chinese test set were translated into English using commercial APIs, and model-generated Chinese answers were likewise translated into English with the same tool. Both the original Chinese pairs (query and answer) and the translated English pairs were then evaluated by LegalEval-Q under identical conditions. This procedure ensured strict comparability across languages and allowed us to test whether the framework produces consistent assessments when applied to English inputs.  

\begin{figure}[h]
    \centering
    \includegraphics[width=0.85\textwidth]{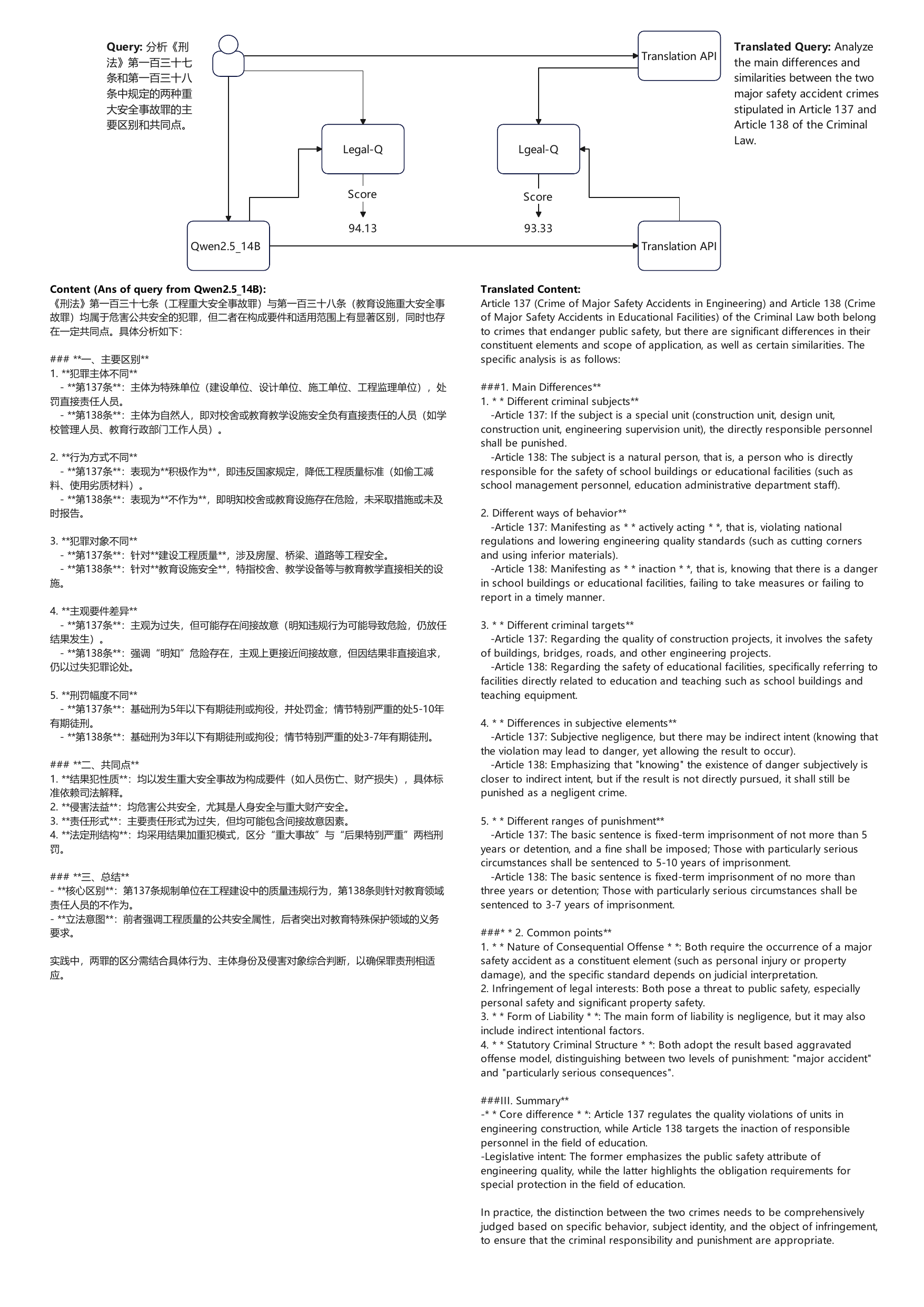}
    \caption{Illustrative workflow of cross-language evaluation in LegalEval-Q.}
    \label{fig:crosslang_workflow}
\end{figure}

As an illustrative example, one representative query was translated into English together with the model-generated Chinese answer. Both the original Chinese pair and the translated English pair were evaluated by LegalEval-Q. The framework produced highly consistent results, with scores of 94.13 and 93.33, respectively. While LegalEval-Q also generates detailed comments and an overall conclusion, for clarity only the evaluation score is shown in Figure~\ref{fig:crosslang_workflow} to highlight the evaluation pipeline and cross-linguistic comparability.  

\subsection{Paired-Sample Statistical Tests}
In the first part of the validation, we directly compared the evaluation scores produced by LegalEval-Q on the original Chinese dataset with those obtained from its English translation. Both the queries and the model-generated answers were translated into English and evaluated under identical conditions.  

The average score for the Chinese dataset was 69.11 (standard deviation 24.45), while the English dataset yielded an average score of 69.40 (standard deviation 23.96). The mean difference was $-0.29$ (standard deviation 7.52), with a range between $-25.23$ and $34.25$.  

A paired t-test indicated no significant difference between the two groups ($t = -0.8821$, $p = 0.3781$, degrees of freedom = 539, 95\% confidence interval: $[-0.92, 0.35]$). The effect size was negligible (Cohen’s $d = -0.038$). Since the Shapiro–Wilk test suggested that the distribution of differences deviated from normality ($W = 0.9600$, $p < 0.001$), we additionally conducted a non-parametric Wilcoxon signed-rank test, which again showed no significant difference ($V = 71,256$, $p = 0.624$).  

\begin{table}[h]
    \centering
    \caption{Results of paired-sample statistical tests comparing Chinese and translated English scores.}
    \label{tab:paired_test}
    \begin{tabular}{lll}
    \toprule
    \textbf{Test Item} & \textbf{Statistic} & \textbf{Value} \\
    \midrule
    \multicolumn{3}{l}{\textbf{1. Basic Statistics}} \\
    Sample size & N & 540 \\
    Original scores mean $\pm$ std & $\mu_1 \pm \sigma_1$ & 69.11 $\pm$ 24.45 \\
    Translated scores mean $\pm$ std & $\mu_2 \pm \sigma_2$ & 69.401 $\pm$ 23.96 \\
    Difference mean $\pm$ std & $\mu_d \pm \sigma_d$ & -0.29 $\pm$ 7.52 \\
    Difference range & [min, max] & [-25.23, 34.25] \\
    \midrule
    \multicolumn{3}{l}{\textbf{2. Paired t-test}} \\
    T-statistic & $t$ & -0.8821 \\
    P-value & $p$ & 0.3781 \\
    Degrees of freedom & df & 539 \\
    95\% CI & CI & (-0.92, 0.35) \\
    Effect size & Cohen's $d$ & -0.038 \\
    \midrule
    \multicolumn{3}{l}{\textbf{3. Normality Test (Shapiro-Wilk)}} \\
    Statistic & $W$ & 0.9600 \\
    P-value & $p$ & $<$0.001 \\
    \midrule
    \multicolumn{3}{l}{\textbf{4. Wilcoxon Signed-Rank Test}} \\
    Statistic & $V$ & 71,256 \\
    P-value & $p$ & 0.624 \\
    \bottomrule
    \end{tabular}
\end{table}

The results in Table~\ref{tab:paired_test} demonstrate that there is no statistically significant difference between the Chinese and English evaluations, and the effect size is negligible. This provides strong evidence that LegalEval-Q produces stable evaluations across languages.

\subsection{Visual Diagnostics}
\begin{figure}[h]
    \centering
    \includegraphics[width=0.85\textwidth]{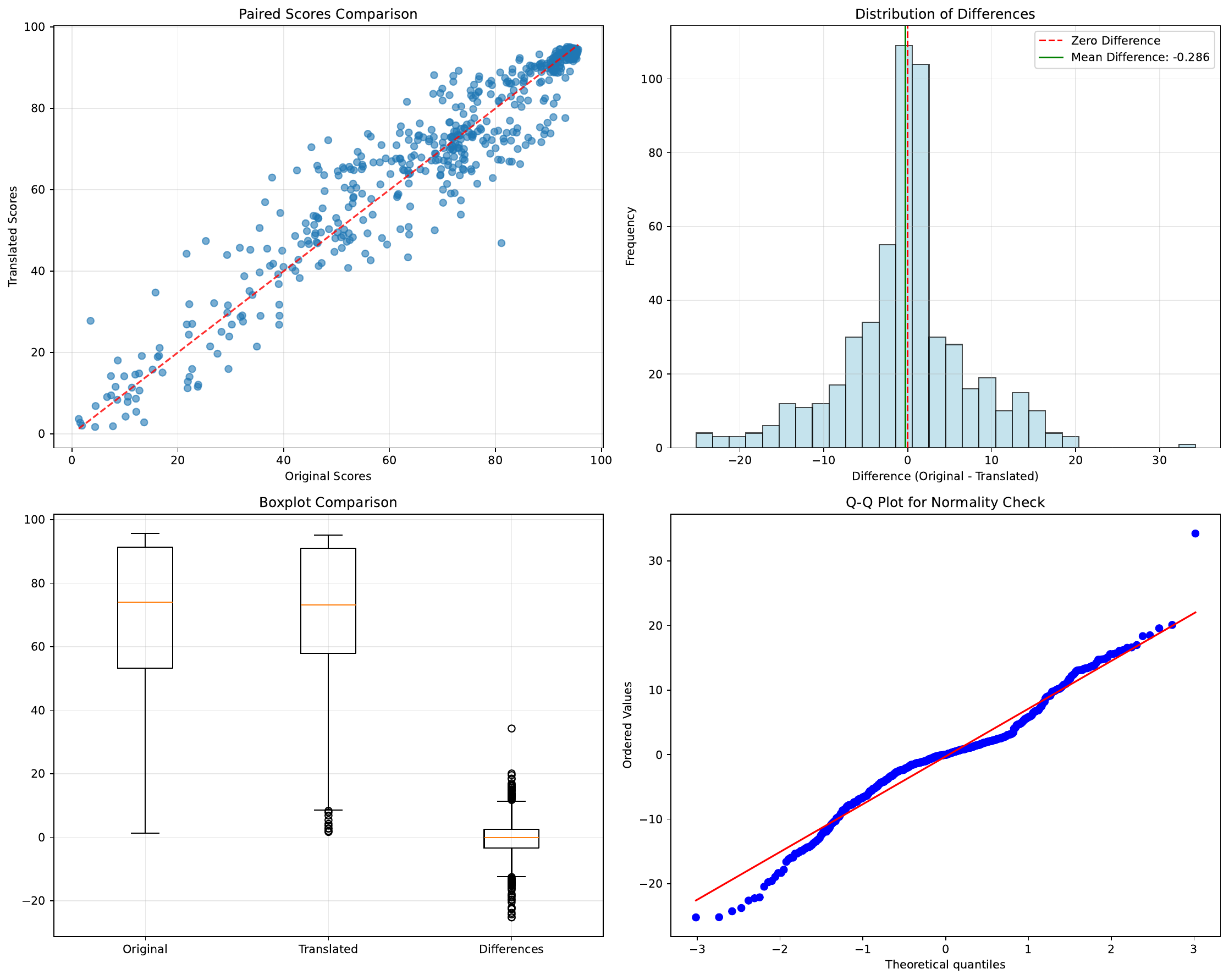}
    \caption{Four-in-one visual diagnostics comparing Chinese and English evaluation scores.}
    \label{fig:crosslang_diagnostics}
\end{figure}

To complement the statistical tests, we examined four diagnostic plots. As shown in Figure~\ref{fig:crosslang_diagnostics}, the scatter plot of paired scores indicated a linear trend with most points clustered near the 45-degree line. The boxplots revealed substantial overlap in distributions, and the residual distribution plot confirmed that deviations were centered around zero. The Q--Q plot showed alignment with the diagonal reference line, with only minor deviations at the tails.  

These diagnostics confirm the statistical results and provide further evidence that the LegalEval-Q framework yields consistent outputs across Chinese and English inputs.

\subsection{Mixed-Effects Model Analysis}
To provide an additional robustness check, we conducted a mixed-effects model analysis, treating \emph{item} as a random effect and \emph{language} (Chinese vs. English) as a fixed effect. As shown in Table~\ref{tab:framework_validity}, the estimated language effect was 0.286 points with a non-significant $p$-value ($p = 0.846$), and the average item-level difference was 0.286 points (range $-4.57$ to $5.88$).  

\begin{table}[htbp]
    \centering
    \caption{Results of mixed-effects model analysis of Chinese vs. English scores.}
    \label{tab:framework_validity}
    \begin{tabular}{lll}
    \toprule
    \textbf{Analysis Component} & \textbf{Statistic} & \textbf{Value} \\
    \midrule
    \multicolumn{3}{l}{\textbf{1. Mixed Effects Model (Language Effect)}} \\
    Fixed Effect (English) & $\beta$ & 0.286 \\
    P-value & $p$ & 0.846 \\
    95\% Confidence Interval & CI & (-2.60, 3.17) \\
    \midrule
    \multicolumn{3}{l}{\textbf{2. Item-level Differences}} \\
    Mean Difference & $\mu$ & 0.286 \\
    Standard Deviation & $\sigma$ & 2.3415 \\
    Range & [min, max] & [-4.567, 5.88] \\
    \bottomrule
    \end{tabular}
\end{table}

The mixed-effects model results further corroborate the paired-sample statistical tests, indicating that language differences do not systematically affect evaluation outcomes.

\subsection{Summary of Findings}
Across all validation methods, including paired-sample statistical tests, diagnostic plots, and mixed-effects modeling, no systematic bias was observed between the original Chinese data and its English translation. Although a slight upward tendency was noted in English scores, this difference was not statistically significant. These findings confirm that the LegalEval-Q framework maintains consistent and reliable performance across languages, providing strong evidence of its preliminary cross-linguistic applicability.

\end{document}